\documentclass[]{fairmeta}

\usepackage[utf8]{inputenc}
\usepackage{microtype}
\usepackage{graphicx}
\usepackage{booktabs}
\usepackage{colortbl}
\usepackage{todonotes}
\usepackage{multirow}
\usepackage{subcaption}
\usepackage{amsmath}
\usepackage{dsfont}
\usepackage{amssymb}
\usepackage{listings}
\usepackage{array}
\usepackage{lipsum}
\usepackage{longtable} 
\usepackage{adjustbox}
\usepackage{makecell}
\usepackage{rotating}
\usepackage{url}
\usepackage{threeparttable}
\usepackage{tabularx}
\usepackage{subcaption}
\newtcblisting{promptbox}[1][]{
    breakable,
    enhanced,
    coltitle=white,
    listing only,
    boxed title style={sharp corners, frame hidden},
    fonttitle=\ttfamily\centering\bfseries,
    title={#1},
    listing options={
        language={},
        basicstyle=\small\ttfamily,
        breaklines=true,
        breakautoindent=false,
        breakindent=0pt,
        columns=fullflexible,
        showstringspaces=false,
    },
    boxsep=2pt,
    top=2pt,
    bottom=2pt,
}

\title{Cycle-Consistent Search: Question Reconstructability as a Proxy Reward for Search Agent Training}

\author[1,2,*]{Sohyun An}
\author[1]{Shuibenyang Yuan}
\author[1]{Hayeon Lee}
\author[2,\dagger]{Cho-Jui Hsieh}
\author[1,\dagger]{Alexander Min}

\affiliation[1]{Meta Superintelligence Labs}
\affiliation[2]{UCLA}

\contribution[*]{Work done at Meta}
\contribution[\dagger]{Joint last author}

\abstract{
Reinforcement Learning (RL) has shown strong potential for optimizing search agents in complex information retrieval tasks. However, existing approaches predominantly rely on gold supervision, such as ground-truth answers, which is difficult to scale. To address this limitation, we propose Cycle-Consistent Search (CCS), a gold-supervision-free framework for training search agents, inspired by cycle-consistency techniques from unsupervised machine translation and image-to-image translation.
Our key hypothesis is that an optimal search trajectory, unlike insufficient or irrelevant ones, serves as a lossless encoding of the question's intent. Consequently, a high-quality trajectory should preserve the information required to accurately reconstruct the original question, thereby inducing a reward signal for policy optimization. However, naive cycle-consistency objectives are vulnerable to information leakage, as reconstruction may rely on superficial lexical cues rather than the underlying search process.
To reduce this effect, we apply information bottlenecks, including exclusion of the final response and named entity recognition (NER) masking of search queries. These constraints force reconstruction to rely on retrieved observations together with the structural scaffold, ensuring that the resulting reward signal reflects informational adequacy rather than linguistic redundancy. Experiments on question-answering benchmarks show that CCS achieves performance comparable to supervised baselines while outperforming prior methods that do not rely on gold supervision. These results suggest that CCS provides a scalable training paradigm for training search agents in settings where gold supervision is unavailable.
}

\date{\today}
\correspondence{Sohyun An at \email{sohyun0423@cs.ucla.edu}}

\begin{document}

\maketitle

\section{Introduction}

Recent advances in Large Language Models (LLMs) \citep{brown2020language, chung2024scaling} have led to the emergence of search agents \citep{search-o1, search-r1, r1-searcher, deepresearcher, ReSearch} that navigate complex information environments through iterative planning and tool use \citep{react}. Unlike conventional single-step retrieval systems, these agents actively formulate search queries, inspect retrieved observations, and adapt subsequent actions to answer complex, multi-faceted questions.
Reinforcement Learning (RL) has become a standard framework for optimizing such sequential decision-making processes \citep{ppo, grpo}. However, existing RL-based search agents typically rely on gold supervision, such as ground-truth answers, to define reward signals. This reliance creates a fundamental scalability bottleneck: in specialized or rapidly evolving domains, such supervision is often prohibitively expensive or unavailable, making it difficult to construct reliable rewards for search trajectories.

\begin{figure*}[!t]
    \centering
    \includegraphics[width=1\linewidth]{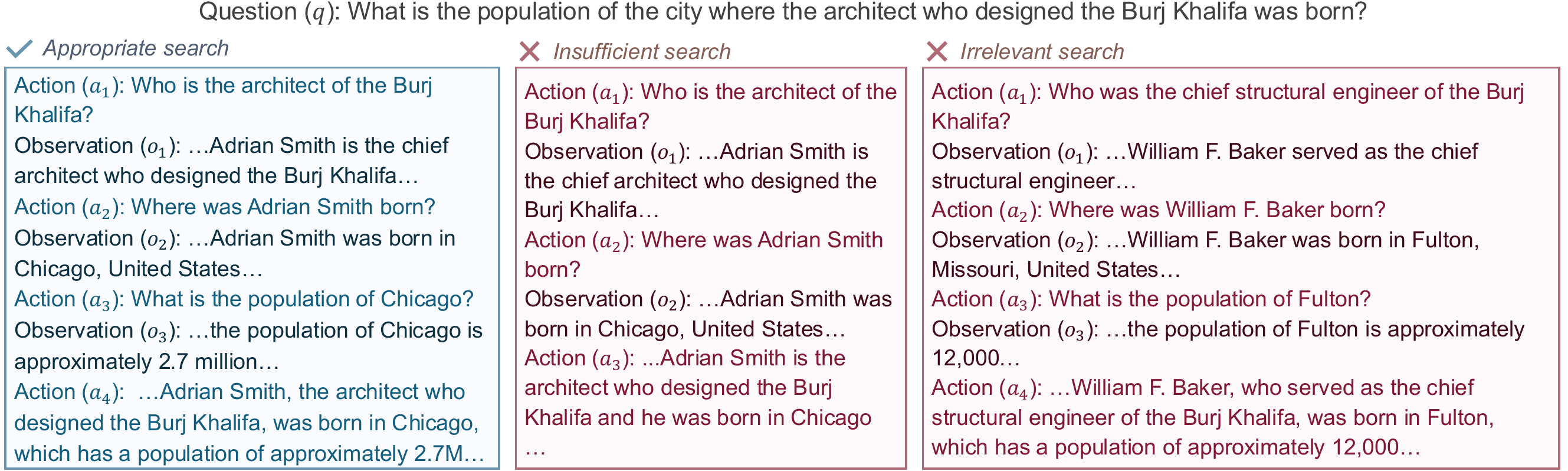}
    \caption{\textbf{Examples of search trajectories.} A high-quality trajectory (left) constitutes a lossless encoding of the question's intent, preserving all necessary information required to reconstruct the original question. In contrast, trajectories that involve insufficient (middle) or irrelevant (right) search steps fail to preserve this information, resulting in incomplete or distorted representations of the question.}
    \label{fig:ccs_ex}
\end{figure*}

To address this limitation, we propose \textbf{Cycle-Consistent Search (CCS)}, a gold-supervision-free framework for training search agents using the internal structure of the search process itself. Our approach is inspired by cycle-consistency techniques from unsupervised machine translation \citep{he2016dual, lample2017unsupervised, han2021unsupervised} and image-to-image translation \citep{cyclegan, liu2017unsupervised, huang2018multimodal}, where the quality of a transformation is assessed by whether the original input can be recovered through an inverse mapping.
We extend this idea to search by treating a search trajectory $\tau$ as an information-preserving encoding of the source question $q$. As illustrated in \Cref{fig:ccs_ex}, trajectories that omit essential steps for multi-hop questions (middle) or drift toward irrelevant information (right) fail to preserve the information needed to recover the original question, whereas high-quality trajectories (left) do. Formally, our central hypothesis is that, under a high-quality policy $\pi$, the source question should be recoverable from the resulting trajectory: $q \xrightarrow{\pi} \tau \xrightarrow{\phi} \hat{q} \approx q$, where $\phi$ denotes a reconstruction function that maps a search trajectory back to the source question. Under this view, reconstruction quality provides a proxy for trajectory quality.

\begin{figure*}[!t]
    \centering
    \includegraphics[width=1\linewidth]{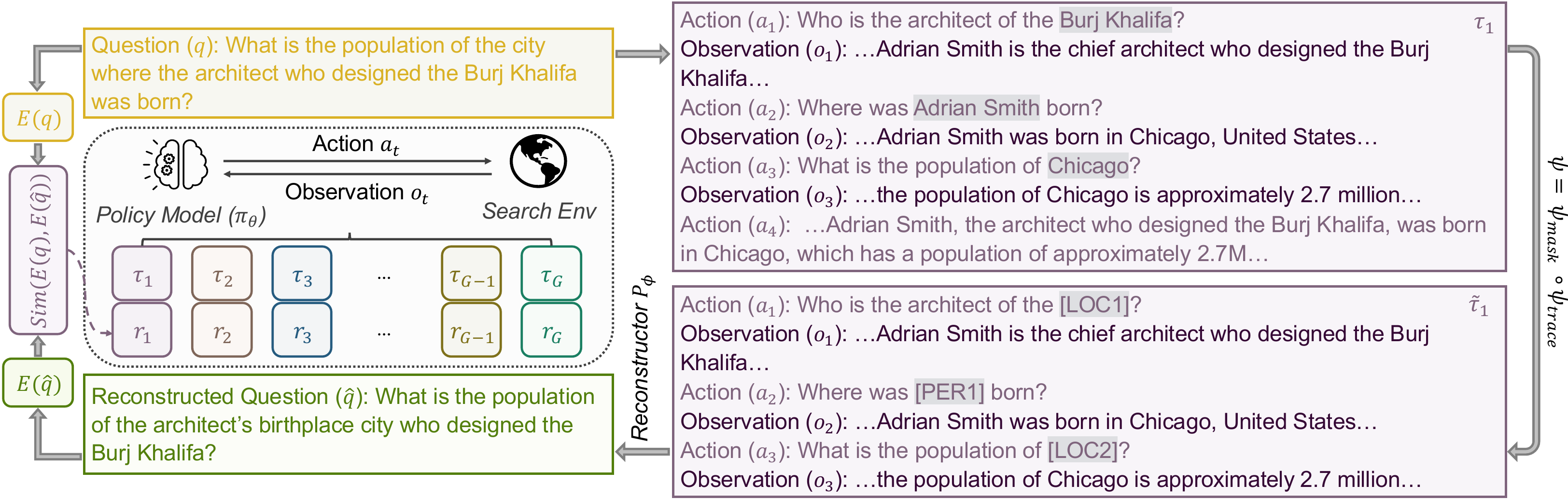}
    \caption{\textbf{Overview of the Cycle-Consistent Search (CCS) framework.} The agent $\pi_\theta$ generates search trajectories $\tau$ to address a question $q$. To ensure the search process is driven by information gain rather than lexical copying, $\tau$ is passed through a strategic information bottleneck ($\psi$) that removes the $a_T$ and masks entities within actions (highlighted in grey). A frozen reconstructor $P_\phi$ then attempts to recover the original question from $\tilde{\tau}$. The agent is optimized using GRPO based on the semantic similarity between $q$ and $\hat{q}$, encouraging trajectories that capture sufficient evidence.}
    \label{fig:ccs_main}
\end{figure*}

A central challenge in applying cycle-consistency to search is \emph{information leakage}. Search trajectories may contain superficial lexical cues that enable reconstruction without reflecting genuine search quality. For example, search queries often overlap with the wording of the original question, and final responses may partially restate the question for context, allowing the original question to be reconstructed even when the underlying search process is uninformative or low quality. To mitigate this issue, we introduce two information bottlenecks: excluding the final response from the trajectory and applying Named Entity Recognition (NER) masking \citep{yamada2020luke, sun2019ernie, liu2020k} to search queries, replacing entities such as names and locations with generic tags (e.g., \texttt{[LOC]}). These constraints reduce direct lexical shortcuts and encourage reconstruction to depend on retrieved observations together with the structural scaffold (\Cref{fig:ccs_main}).

Under this formulation, training a search agent amounts to enforcing cycle-consistency over search trajectories, with reconstruction quality on an information-bottlenecked trajectory $\tilde{\tau}$ serving as a proxy reward for trajectory quality. We evaluate CCS on seven representative question-answering benchmarks and a deep research benchmark. The results show that CCS achieves performance comparable to supervised baselines while outperforming existing methods that do not rely on gold supervision. These findings suggest that cycle-consistency can provide a scalable proxy objective for training search agents in settings where gold supervision is unavailable.

In summary, our contributions are as follows:
\begin{itemize}
    \item We introduce Cycle-Consistent Search (CCS), a gold-supervision-free framework for training search agents by enforcing cycle-consistency between questions and search trajectories.
    \item We identify information leakage as a practical challenge in cycle-consistent search training and use information bottlenecks to reduce lexical shortcuts.
    \item We show empirically, across seven question-answering benchmarks and a deep research benchmark, that CCS achieves performance competitive with supervised methods while outperforming prior approaches that do not rely on gold supervision.
\end{itemize}
\section{Related Works}

\paragraph{From Multi-Step Retrieval to Interactive Search Agents}
Prior work in information retrieval has largely centered on single-shot retrieve-and-read pipelines \citep{lee-etal-2019-latent, guu2020retrieval, karpukhin-etal-2020-dense}. This paradigm expanded with the emergence of LLM-based systems that perform retrieval over multiple steps. Prompted tool-use frameworks such as ReAct \citep{react} and multi-step retrieval methods such as IRCoT \citep{ircot} showed that LLMs can improve retrieval coverage on complex queries by iteratively issuing retrieval actions conditioned on intermediate reasoning or generations. These approaches established the broader view that retrieval can be formulated as a sequential, interactive process rather than a single retrieval step.
Building on this perspective, subsequent work increasingly developed interactive search agents that actively generate queries, consume observations, and iteratively refine their information needs over the course of search. Search-o1 \citep{search-o1} demonstrated the effectiveness of this search-centric agent design through prompting-based iterative query formulation and evidence collection over the web. More recent approaches, including \citet{search-r1}, \citet{r1-searcher}, \citet{deepresearcher}, and \citet{ReSearch}, have further advanced this paradigm by training search agents with reinforcement learning \citep{ppo, grpo}. In these approaches, search is explicitly modeled as sequential decision-making over query actions and observation consumption, making RL a natural framework for optimizing end-to-end search behavior.

\paragraph{Beyond Gold Supervision for Search Agents}
Many RL-based search agents have demonstrated strong performance, but they remain dependent on gold supervision, such as ground-truth answers, to construct reliable reward signals. This dependence creates a scalability bottleneck in domains where high-quality annotations are scarce, expensive, or rapidly evolving.
To reduce the cost of annotation, several recent methods have explored alternatives to gold supervision by leveraging alternative supervision signals, including intrinsic signals and self-generated pseudo-labels. RLIF \citep{rlif}, for example, derives reward signals from the model's internal confidence to guide training without external labels.
Rubric-based LLM judges \citep{geval}, which we refer to as Constitutional Judges, can also be used to provide reward signals by scoring model outputs according to a predefined set of evaluation criteria.
Test-time RL methods such as TTRL \citep{ttrl} instead use agreement among multiple sampled rollouts as a proxy signal for updating policies on unlabeled instances. While these approaches reduce reliance on human annotation, their training signals are not specifically aligned with the central objective of search: rewarding trajectories that acquire sufficient and necessary external evidence to satisfy the original information need.

\paragraph{Extending Cycle-Consistency to Search Trajectories}
Cycle-consistency has been widely used as a learning principle that encourages transformations to preserve essential information, notably in unsupervised machine translation \citep{he2016dual, lample2017unsupervised, han2021unsupervised} and image-to-image translation \citep{cyclegan, liu2017unsupervised, huang2018multimodal}. Its central intuition is that a good transformation should retain enough information for the original input to be recoverable through an inverse mapping. Related reconstruction-based objectives have also been explored in retrieval-augmented pretraining, where retrieved documents serve as latent variables for reconstructing targets \citep{guu2020retrieval}. In the search setting, we consider cycle-consistency over agentic search trajectories, treating a trajectory as an information-preserving encoding of the source question. A central challenge in this setting is information leakage: lexical overlap between the question and intermediate search queries (or the final response) can enable trivial reconstruction without reflecting genuine search quality. We address this issue through information bottlenecks, so that successful reconstruction must rely on retrieved observations together with the structural scaffold rather than surface-form redundancy.


\section{Method}
In this section, we formally introduce \textbf{Cycle-Consistent Search (CCS)}, a framework designed to train search agents \emph{without gold supervision}. We first establish the theoretical connection between search trajectory optimization and cycle-consistency. We then derive an information-theoretic objective and introduce structural constraints that prevent information leakage. Finally, we detail the optimization procedure using Group Relative Policy Optimization (GRPO). A schematic overview of CCS is shown in \Cref{fig:ccs_main}.

\subsection{Cycle-Consistency as Information Conservation}
\label{sec:cycle-consistency}
Our approach is grounded in the principle of cycle-consistency, widely utilized in unsupervised machine translation \citep{he2016dual, lample2017unsupervised, han2021unsupervised} and unpaired image-to-image translation \citep{cyclegan, liu2017unsupervised, huang2018multimodal}. Let $\mathcal{X}$ and $\mathcal{Y}$ denote two distinct domains. The fundamental assumption in these settings is that an ideal mapping $F: \mathcal{X}\rightarrow \mathcal{Y}$ preserves the essential semantic content of the input, such that there exists an inverse mapping $G:\mathcal{Y}\rightarrow \mathcal{X}$ satisfying $x \approx G(F(x))$ for any $x\in\mathcal{X}$. This implies that the transformation is information-preserving: the latent information required to reconstruct $x$ is conserved throughout the transformation.

We transpose this principle to agentic search. A user's question $q\in\mathcal{Q}$ represents a semantic intent, and a search trajectory $\tau\in\mathcal{T}$ represents a dynamic expansion of that intent into a sequence of actions and observations. Under the assumption of information conservation, an optimal search trajectory $\tau^*$ acts as a \emph{lossless encoding} of the question $q$. Consequently, if $\tau$ contains the necessary and sufficient evidence to resolve $q$, it should be possible to reconstruct $q$ solely from $\tau$. Formally, this establishes a cycle $q \rightarrow \tau \rightarrow \hat{q}$, where the closeness between $q$ and $\hat{q}$ serves as a proxy for the quality of the intermediate representation $\tau$, independent of external supervision.

\subsection{Problem Formulation and Objective}
\label{sec:problem_formulation}
We model the search process as a Partially Observable Markov Decision Process (POMDP). The agent, parameterized by $\theta$, is defined as a policy $\pi_\theta(\cdot \mid q, h_{t-1})$, where $h_{t-1}$ is the history of past actions and observations. At each intermediate step $t<T$, the agent issues an \emph{action} (search query) $a_t$ and receives an \emph{observation} (search results) $o_t$ from the search environment. After $T-1$ search steps, the agent produces a \emph{final response} $a_T$. The full trajectory is $\tau = (a_1, o_1, \ldots, a_{T-1}, o_{T-1}, a_T)$.

Under the cycle-consistency view in \Cref{sec:cycle-consistency}, the training objective can be formulated as maximizing the mutual information (MI) between the original question $q$ and the generated trajectory $\tau$.
Here, the mutual information $I(Q; \mathcal{T}_\theta)$ is defined as:
\begin{equation}
    I(Q; \mathcal{T}_\theta) = H(Q) - H(Q \mid \mathcal{T}_\theta).
\end{equation}
Since the entropy of the question distribution $H(Q)$ is constant with respect to $\theta$, maximizing MI is equivalent to minimizing the conditional entropy $H(Q \mid \mathcal{T}_\theta)$, which corresponds to maximizing the expected log-likelihood of the question given the trajectory:
\begin{equation}
    \mathcal{J}(\theta) = \mathbb{E}_{q \sim \mathcal{D}, \tau \sim \pi_\theta(\cdot\mid q)} \left[ \log P(q \mid \tau) \right].
\end{equation}
Directly modeling the true posterior $P(q \mid \tau)$ is intractable. We therefore employ a variational surrogate using a fixed, pre-trained reconstructor $P_\phi(q \mid \cdot)$, which serves as a tractable approximation to the inverse mapping $G$.

However, naively maximizing $\log P_\phi(q\mid\tau)$ can lead to degenerate solutions in which the agent encodes $q$ into the action sequence through trivial lexical copying (e.g., by repeating the question in a search query or partially restating it in the final response for context), allowing the reconstructor to recover $q$ without relying on the search results. To mitigate this issue, we introduce a bottleneck transformation $\psi: \mathcal{T} \rightarrow \tilde{\mathcal{T}}$ and optimize the following objective:
\begin{equation}
    \theta^* = \operatorname*{argmax}_\theta
    \mathbb{E}_{q \sim \mathcal{D},\, \tau \sim \pi_\theta(\cdot\mid q)}
    \left[ \log P_\phi\big(q \mid \psi(\tau)\big) \right].
\end{equation}

\subsection{Strategic Information Bottlenecks}
The transformation function $\psi$ imposes structural constraints on the trajectory so that it retains the search scaffold while reducing lexical shortcuts that would otherwise allow trivial reconstruction of the source question.
We define $\psi$ as the composition of two operations: \emph{final response exclusion} and \emph{entity masking}: $\psi = \psi_{\text{mask}} \circ \psi_{\text{trace}}$.

\paragraph{Final response exclusion.}
We first address leakage through the final response. In standard QA interactions, the final response $a_T$ often paraphrases $q$ to provide context. Including it in the reconstruction input can bypass the evaluation of the search process. Therefore, we define the trace operator $\psi_{\text{trace}}$ which truncates the trajectory to exclude the final generation step, preserving only the investigative actions and observations: $\psi_{\text{trace}}(\tau) = (a_1, o_1, \ldots, a_{T-1}, o_{T-1})$.

\paragraph{Entity masking on search actions.}
We next mitigate leakage through search queries. Raw search queries $a_t$ typically exhibit high lexical overlap with $q$. To prevent the reconstructor from recovering $q$ from these surface-level cues even when the underlying search process is uninformative, we apply a masking operator $\psi_{\text{mask}}$ that identifies named entities within each search query $a_t$ (e.g., ``Burj Khalifa'') and replaces them with generic typed tags (e.g., \texttt{[LOC]}), yielding a masked action $\tilde{a}_t$. The observations $o_t$ remain unmasked. This operation substantially reduces the direct lexical channel from actions to the question, forcing reconstruction to resolve masked content primarily from evidence in $o_t$ rather than from copied words in $a_t$.

The full bottleneck transformation produces a processed trajectory: $\psi(\tau) = \tilde{\tau} = (\tilde{a}_1, o_1, \ldots, \tilde{a}_{T-1}, o_{T-1})$.
Consequently, to recover the original question from $\tilde{\tau}$, the reconstructor must leverage the information in the observations together with the structural intent expressed by the masked queries (e.g., $\tilde{\tau}_1$ in \Cref{fig:ccs_main}).
Trajectories with an incorrect or insufficient search scaffold, or with irrelevant or insufficient observations, will fail to preserve the relational structure or informational content required to reconstruct the original question, resulting in low reconstruction fidelity and thus low reward.
By tying the agent's incentive to both the correctness of the search scaffold and the informativeness of the resulting observations, CCS makes the reward sensitive to the quality of the search trajectory. In this way, the cycle-consistency objective penalizes trajectories that are incorrect, incomplete, or noisy, while favoring those that preserve the relational structure of the question and gather the information required for faithful reconstruction.

\subsection{Optimization via Group Relative Policy Optimization}
The objective in \Cref{sec:problem_formulation} frames search-agent training as maximizing the reconstructability of the original question from the bottlenecked trajectory. This naturally defines a trajectory-level reward: trajectories that better preserve the information needed to reconstruct $q$ receive higher rewards.
We optimize the search agent with reinforcement learning using our cycle-consistency reward. Specifically, we adopt Group Relative Policy Optimization (GRPO) \citep{grpo}, which has proven effective for search-agent optimization \citep{search-r1} and avoids the need for a separate value-function critic by using group-based relative advantages.

The RL objective is to optimize $\pi_\theta$ by maximizing expected returns over questions $q\sim\mathcal{D}$:
\begin{equation}
    \max_{\pi_\theta} \mathbb{E}_{q \sim \mathcal{D}, \tau \sim \pi_\theta(\cdot\mid q)} \big[r(\tau)\big].
\end{equation}
For each question $q$, the agent samples a group of $G$ trajectories $\{\tau_1,\tau_2,\dots,\tau_G\}$ from the current policy $\pi_{\theta_{\text{old}}}$. For each $\tau_i$, we compute a reconstructed question $\hat{q}_i \sim P_\phi(\cdot \mid \psi(\tau_i))$ and define the reward as the semantic similarity between $q$ and $\hat{q}_i$:
\begin{equation}
    r(\tau_i) = \mathrm{Sim}\big(E(q), E(\hat{q}_i)\big),
    \label{eq:embedding_sim}
\end{equation}
where $E(\cdot)$ is a sentence embedding function and $\mathrm{Sim}(\cdot,\cdot)$ denotes cosine similarity.

We then compute an advantage $A_i$ by normalizing rewards within the group:
\begin{equation}
    A_i = \frac{r(\tau_i) - \mu(\{r(\tau_j)\}_{j=1}^G)}{\sigma(\{r(\tau_j)\}_{j=1}^G) + \epsilon},
\end{equation}
where $\mu$ and $\sigma$ denote the mean and standard deviation of group rewards. The GRPO objective is then:
\begin{equation}
\begin{aligned}
\mathcal{L}_{GRPO}(\theta)
= \mathbb{E}_{q \sim \mathcal{D}} \Bigg[
\frac{1}{G} \sum_{i=1}^G
\Big(
\min \Big(
\frac{\pi_\theta(\tau_i|q)}{\pi_{\theta_{old}}(\tau_i|q)} A_i,
\text{clip}\Big(
\frac{\pi_\theta(\tau_i|q)}{\pi_{\theta_{old}}(\tau_i|q)},
1-\epsilon, 1+\epsilon
\Big) A_i
\Big)
\\
- \beta \mathbb{D}_{KL}(\pi_\theta \,\|\, \pi_{ref})
\Big)
\Bigg].
\end{aligned}
\end{equation}
Here, observations are treated as environment outputs and $\pi_\theta(\tau_i\mid q)$ refers to the likelihood of the agent-generated tokens/actions \citep{search-r1}.
This objective increases the likelihood of trajectories that yield higher reconstruction fidelity relative to the group average, thereby encouraging search behaviors that better preserve the structure and information needed to recover the original question.
\section{Experiments}
\label{sec:experiments}
\begin{table}[htbp]
\centering
\caption{\textbf{Main results.} The best and second-best performances are shown in bold and underline, respectively. $^{\dagger}$/$^{\star}$ denotes in-domain/out-of-domain datasets.}
\resizebox{\textwidth}{!}{
\begin{tabular}{lcccccccc}
\toprule
\multirow{2}{*}{\textbf{Methods}} & \multicolumn{3}{c}{\textbf{General QA}} & \multicolumn{4}{c}{\textbf{Multi-Hop QA}} & \multirow{2}{*}{\textbf{Avg.}} \\
\cmidrule(lr){2-4} \cmidrule(lr){5-8}
 & \textbf{NQ}$^\dagger$ & \textbf{TriviaQA}$^*$ & \textbf{PopQA}$^*$ & \textbf{HotpotQA}$^\dagger$ & \textbf{2wiki}$^*$ & \textbf{Musique}$^*$ & \textbf{Bamboogle}$^*$ & \\
\midrule
\rowcolor{gray!10}
\multicolumn{9}{c}{\textbf{\textit{Qwen2.5-7B-Instruct}}} \\
\textbf{Model Inference} & & & & & & & & \\
\hspace{1em}Direct Inference & 0.344 & 0.111 & 0.144 & 0.279 & 0.248 & 0.091 & 0.120 & 0.191 \\
\hspace{1em}CoT & 0.430 & 0.139 & 0.203 & 0.325 & 0.318 & 0.128 & 0.384 & 0.275 \\
\hspace{1em}IRCoT & 0.608 & 0.761 & 0.515 & 0.486 & 0.279 & 0.129 & 0.304 & 0.440 \\
\hspace{1em}RAG & 0.645 & 0.805 & 0.535 & 0.501 & 0.331 & 0.135 & 0.304 & 0.465 \\
\hspace{1em}Search-o1 & 0.654 & 0.588 & 0.462 & 0.486 & 0.481 & 0.197 & 0.536 & 0.486 \\
\textbf{W. Gold Supervision} & & & & & & & & \\
\hspace{1em}SFT & 0.366 & 0.587 & 0.220 & 0.312 & 0.267 & 0.087 & 0.224 & 0.295 \\
\hspace{1em}Search-R1 (EM) & 0.693 & \underline{0.872} & 0.627 & \underline{0.587} & \textbf{0.617} & \textbf{0.244} & \underline{0.568} & \underline{0.601} \\
\textbf{W.O. Gold Supervision} & & & & & & & & \\
\hspace{1em}TTRL & 0.695 & 0.500 & 0.593 & 0.500 & 0.455 & 0.173 & 0.540 & 0.544 \\
\hspace{1em}CJ & \underline{0.704} & 0.865 & \underline{0.633} & 0.545 & 0.560 & 0.209 & 0.544 & 0.580 \\
\hspace{1em}RLIF & 0.672 & 0.846 & 0.586 & 0.536 & 0.444 & 0.190 & 0.408 & 0.526 \\
\hspace{1em}CCS & \textbf{0.712} & \textbf{0.880} & \textbf{0.639} & \textbf{0.598} & \underline{0.567} & \underline{0.239} & \textbf{0.608} & \textbf{0.606} \\
\midrule
\rowcolor{gray!10}
\multicolumn{9}{c}{\textbf{\textit{Qwen3-4B-Instruct-2507}}} \\
\textbf{Model Inference} & & & & & & & & \\
\hspace{1em}Direct Inference & 0.349 & 0.091 & 0.119 & 0.260 & 0.253 & 0.101 & 0.256 & 0.204 \\
\hspace{1em}CoT & 0.404 & 0.142 & 0.204 & 0.294 & 0.260 & 0.114 & 0.408 & 0.261 \\
\hspace{1em}IRCoT & 0.613 & 0.770 & 0.525 & 0.457 & 0.203 & 0.131 & 0.264 & 0.423 \\
\hspace{1em}RAG & 0.688 & 0.803 & 0.564 & 0.472 & 0.264 & 0.126 & 0.312 & 0.461 \\
\hspace{1em}Search-o1 & 0.541 & 0.489 & 0.234 & 0.413 & 0.335 & 0.150 & 0.384 & 0.364 \\
\textbf{W. Gold Supervision} & & & & & & & & \\
\hspace{1em}SFT & 0.312 & 0.474 & 0.196 & 0.275 & 0.275 & 0.078 & 0.120 & 0.247 \\
\hspace{1em}Search-R1 (EM) & \textbf{0.760} & \textbf{0.921} & 0.671 & \textbf{0.638} & \textbf{0.668} & \textbf{0.277} & \underline{0.608} & \textbf{0.649} \\
\textbf{W.O. Gold Supervision} & & & & & & & & \\
\hspace{1em}TTRL & 0.740 & \underline{0.902} & 0.630 & 0.540 & 0.488 & 0.018 & 0.482 & 0.543 \\
\hspace{1em}CJ & 0.709 & 0.893 & \underline{0.677} & 0.539 & 0.495 & 0.189 & 0.552 & 0.579 \\
\hspace{1em}RLIF & 0.686 & 0.879 & 0.630 & 0.566 & 0.514 & 0.221 & 0.520 & 0.574 \\
\hspace{1em}CCS & \underline{0.744} & \underline{0.902} & \textbf{0.683} & \underline{0.625} & \underline{0.606} & \underline{0.262} & \textbf{0.632} & \underline{0.636} \\
\midrule
\rowcolor{gray!10}
\multicolumn{9}{c}{\textbf{\textit{Qwen3-32B}}} \\
\textbf{Model Inference} & & & & & & & & \\
\hspace{1em}Direct Inference & 0.485 & 0.308 & 0.277 & 0.368 & 0.340 & 0.144 & 0.224 & 0.307 \\
\hspace{1em}CoT & 0.488 & 0.322 & 0.282 & 0.364 & 0.338 & 0.144 & 0.336 & 0.325 \\
\hspace{1em}IRCoT & 0.522 & 0.697 & 0.443 & 0.411 & 0.223 & 0.113 & 0.312 & 0.389 \\
\hspace{1em}RAG & 0.714 & 0.838 & 0.593 & 0.560 & 0.358 & 0.173 & 0.400 & 0.519 \\
\hspace{1em}Search-o1 & 0.722 & 0.881 & 0.608 & 0.579 & 0.629 & 0.247 & 0.584 & 0.607 \\
\textbf{W. Gold Supervision} & & & & & & & & \\
\hspace{1em}SFT & 0.454 & 0.670 & 0.262 & 0.362 & 0.314 & 0.128 & 0.288 & 0.354 \\
\hspace{1em}Search-R1 (EM) & 0.681 & 0.880 & 0.578 & 0.648 & \underline{0.736} & 0.278 & \textbf{0.742} & \underline{0.649} \\
\textbf{W.O. Gold Supervision} & & & & & & & & \\
\hspace{1em}TTRL & 0.640 & 0.870 & 0.616 & 0.523 & 0.653 & 0.238 & 0.540 & 0.583 \\
\hspace{1em}CJ & \textbf{0.733} & 0.556 & \textbf{0.645} & \underline{0.665} & \textbf{0.739} & \textbf{0.301} & \underline{0.731} & 0.624 \\
\hspace{1em}RLIF & 0.714 & \underline{0.887} & 0.601 & 0.568 & 0.552 & 0.221 & 0.528 & 0.582 \\
\hspace{1em}CCS & \textbf{0.733} & \textbf{0.896} & \underline{0.624} & \textbf{0.666} & 0.718 & \underline{0.284} & 0.712 & \textbf{0.662} \\
\bottomrule
\end{tabular}
}
\label{tbl:main}
\end{table}
We empirically evaluate Cycle-Consistent Search (CCS) across a diverse range of knowledge-intensive tasks. The experimental design is structured to assess the framework's efficacy in both multi-hop and general question-answering scenarios against established baselines.

\subsection{Datasets and Models} 
To ensure a comprehensive assessment, we categorize the evaluation benchmarks into two groups based on search complexity, following \citet{search-r1}. For multi-hop reasoning, which necessitates iterative information retrieval and synthesis, we employ HotpotQA \citep{hotpotqa}, 2WikiMQA \citep{2wikimqa}, MuSiQue \citep{musique}, and Bamboogle \citep{bamboogle}. For general question answering, we utilize Natural Questions (NQ) \citep{nq}, TriviaQA \citep{triviaqa}, and PopQA \citep{popqa}.
We select policy models from the Qwen family \citep{yang2025qwen3} to evaluate performance across different parameter scales and both base and instruct variants: Qwen2.5-7B-Instruct, Qwen3-4B-Instruct-2507, and Qwen3-32B. All evaluations are conducted using Gemini 2.5 Flash \citep{comanici2025gemini} as the evaluator model to ensure consistent assessment of answer accuracy (see \Cref{appendix:evaluator_prompt}).

\subsection{Baselines} 
We compare CCS against a diverse set of baselines spanning model inference-only methods, methods trained with gold supervision, and methods that do not rely on gold supervision.

\textbf{Model Inference:} We include Direct inference, zero-shot Chain-of-Thought (CoT) \citep{cot}, Interleaved Retrieval CoT (IRCoT) \citep{ircot}, Retrieval-Augmented Generation (RAG) \citep{rag}, and Search-o1 \citep{search-o1}, which represent prompting-based or retrieval-based approaches without additional training.

\textbf{With Gold Supervision:} We include Supervised Fine-Tuning (SFT), which trains the model to generate ground-truth answers, and Search-R1 \citep{search-r1}, which optimizes search agents with reinforcement learning using rewards derived from ground-truth answers.

\textbf{Without Gold Supervision:} We compare against RLIF \citep{rlif}, which derives reward signals from the model's internal confidence; Constitutional Judge (CJ) \citep{geval}, which derives reward signals from a rubric-based LLM judge using predefined evaluation criteria; and TTRL \citep{ttrl}, which uses agreement among multiple sampled rollouts as a proxy reward signal.

\subsection{Setup} 
Our experimental protocol is informed by the configuration used in Search-R1 \citep{search-r1}. We utilize the same search engine interface for all methods employing tool use. For the RAG baseline, we employ E5 \citep{e5} as the dense retriever, indexing a 2018 Wikipedia dump \citep{wikidump} as the knowledge source. The training corpus consists of a mixture of the NQ and HotpotQA training sets. Agents are trained for 300 steps with a global batch size of 512. Please refer to \Cref{appendix:implementaion_details} for more details.

\subsection{Results}
We present the main results in \Cref{tbl:main}. Across all three models, CCS achieves the best average performance among methods that do not rely on gold supervision, consistently outperforming both model inference baselines and prior gold-free training methods. In particular, CCS improves over the strongest competing non-gold baseline by 4.5\%, 9.8\%, and 6.1\% on Qwen2.5-7B-Instruct, Qwen3-4B-Instruct-2507, and Qwen3-32B, respectively. These results indicate that cycle-consistency provides an effective training signal for search-agent optimization without access to ground-truth answers.
Notably, CCS is also competitive with gold-supervised training. On Qwen2.5-7B-Instruct and Qwen3-32B, CCS achieves the best overall average performance in the table, surpassing Search-R1 by 0.5 and 1.3 points, respectively. On Qwen3-4B-Instruct-2507, Search-R1 achieves the highest average score, while CCS also delivers strong overall performance. At the dataset level, the strongest method varies across benchmarks, but CCS remains consistently strong across both general QA and multi-hop QA, yielding the most robust overall performance among gold-free methods. Overall, these results show that CCS substantially narrows---and in some settings reverses---the gap between gold-free and gold-supervised search-agent training.
\begin{table*}[t]
\centering
\caption{\textbf{Ablation studies on the components of CCS.} The best and second-best performances are shown in bold and underline, respectively.}
\resizebox{\textwidth}{!}{
\begin{tabular}{lcccccccc}
\toprule
\multirow{2}{*}{\textbf{Methods}} & \multicolumn{3}{c}{\textbf{General QA}} & \multicolumn{4}{c}{\textbf{Multi-Hop QA}} & \multirow{2}{*}{\textbf{Avg.}} \\
\cmidrule(lr){2-4} \cmidrule(lr){5-8}
 & \textbf{NQ}$^\dagger$ & \textbf{TriviaQA}$^*$ & \textbf{PopQA}$^*$ & \textbf{HotpotQA}$^\dagger$ & \textbf{2wiki}$^*$ & \textbf{Musique}$^*$ & \textbf{Bamboogle}$^*$ & \\
\midrule
Actions + Observations + Final Response & \underline{0.700} & 0.852 & \underline{0.618} & 0.543 & 0.476 & 0.202 & \underline{0.536} & 0.561 \\
Actions + Observations  & 0.679 & 0.852 & 0.598 & 0.526 & 0.474 & 0.201 & 0.488 & 0.545 \\
Observations & 0.691 & \underline{0.866} & 0.610 & \underline{0.589} & \textbf{0.588} & \underline{0.225} & 0.520 & \underline{0.584} \\
Masked Actions + Observations  & \textbf{0.712} & \textbf{0.880} & \textbf{0.639} & \textbf{0.598} & \underline{0.567} & \textbf{0.239} & \textbf{0.608} & \textbf{0.606} \\
\bottomrule
\end{tabular}
}
\label{tbl:ablation}
\end{table*}
\begin{figure*}[!t]
    \centering
    \includegraphics[width=1\linewidth]{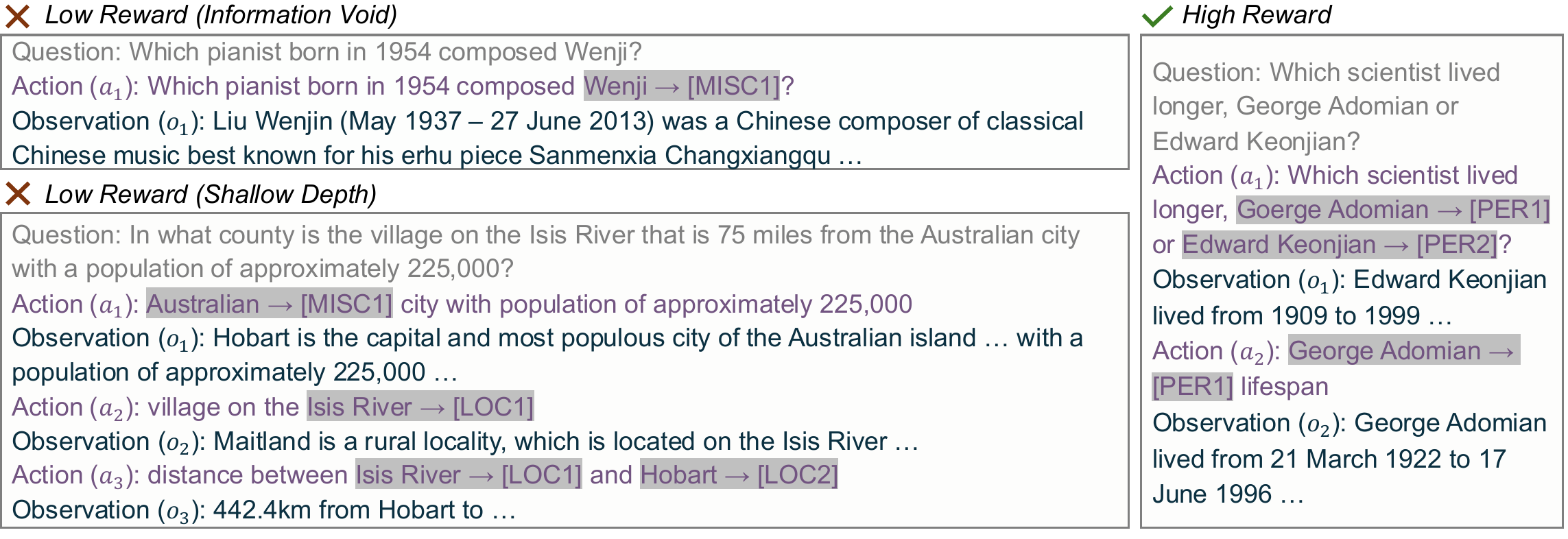}
    \caption{\textbf{Qualitative analysis of search trajectories.} CCS assigns low rewards to trajectories exhibiting (a) Information Void, where the retrieved observations fail to satisfy key entity constraints, and (b) Shallow Depth, where the agent fails to complete the full multi-hop search scaffold. By contrast, high rewards are assigned only to trajectories that preserve both the search structure and the supporting observations needed to faithfully reconstruct the original question. Masked spans are highlighted in gray.}
    \label{fig:qualitative_analysis}
\end{figure*}

\section{Analysis}

\subsection{Ablation Study}
We conduct an ablation study to evaluate the role of the proposed information bottlenecks. As shown in \Cref{tbl:ablation}, variants that expose the reconstructor to unfiltered lexical cues---either by including the final response or by using unmasked actions---perform worse than the CCS bottleneck design (Masked Actions + Observations), with average scores of 0.561 and 0.545, respectively. This suggests that both the final response and raw search queries introduce shortcut signals that weaken the intended training objective.
Using observations alone yields an average score of 0.584, which remains below the 0.606 achieved by the CCS bottleneck design. This performance gap indicates that masked actions contribute more than simply removing lexical leakage: they preserve useful structural intent that helps the reconstructor interpret the observations within a specific investigative context. Overall, these results show that the proposed bottlenecks are effective not only because they suppress surface-level shortcuts, but also because they retain the structural signals needed for faithful reconstruction.

\subsection{Qualitative Investigation}
Qualitative analysis in \Cref{fig:qualitative_analysis} further illustrates how CCS assigns rewards based on the quality of the search trajectory. In the \emph{Information Void} case, the agent retrieves observations that are lexically related to the question (``Liu \texttt{Wenjin}'') but fail to match its key entity constraints (``composed \texttt{Wenji}''). As a result, the trajectory does not preserve the information needed to recover the original question, leading to low reconstruction fidelity and thus a low reward.
In the \emph{Shallow Depth} case, the agent follows an incomplete search scaffold and fails to complete the full multi-hop chain required by the question. Although it identifies an intermediate entity (e.g., ``Hobart''), the trajectory does not gather the additional results needed to resolve the remaining constraints (e.g., the ``county'' in question). Consequently, the reconstructor cannot recover the full relational structure of the original question, and the trajectory receives a low reward. Together, these examples show that CCS rewards trajectories only when they preserve both the search structure and the supporting observations needed for faithful reconstruction.

\begin{figure*}[!t]
    \centering
    \includegraphics[width=1\linewidth]{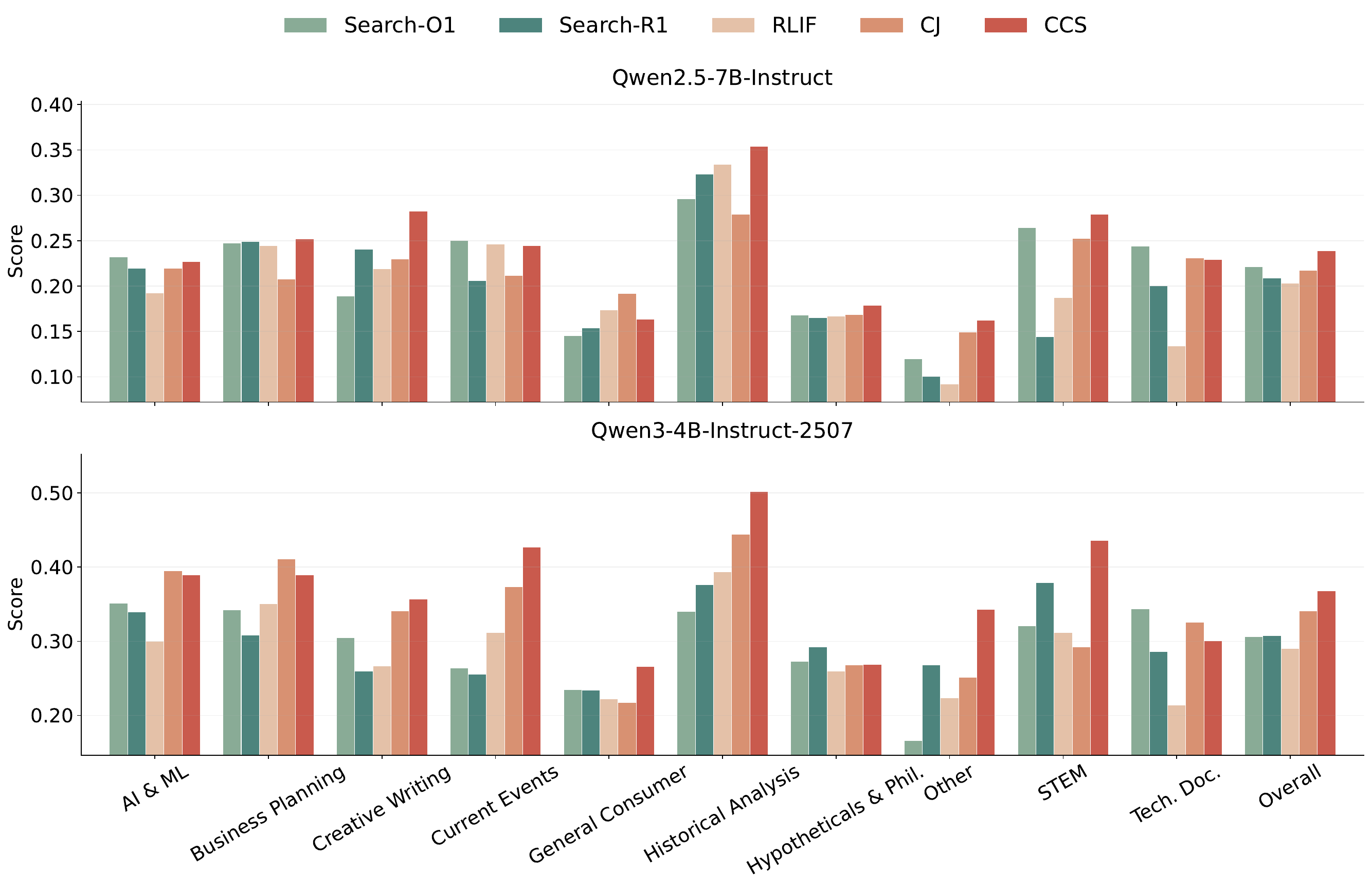}
    \caption{\textbf{Performance comparison on the ResearchRubrics \citep{researchrubrics} benchmark for open-ended Deep Research tasks.} Scores are evaluated using the fine-grained rubric-based framework of \citet{researchrubrics}, with Gemini 2.5 Pro as the judge, and are reported as domain-wise scores and the overall average across ten domains.}
    \label{fig:researchrubrics}
\end{figure*}
\subsection{Open-ended Deep Research Task}
Beyond traditional closed-ended question answering tasks, as reported in \Cref{sec:experiments}, we further evaluate CCS on the open-ended Deep Research setting using the ResearchRubrics benchmark \citep{researchrubrics}. This benchmark requires models to integrate multiple capabilities, including the generation of evidence-backed, long-form responses to open-ended queries, and we conduct experiments with Qwen2.5-7B-Instruct and Qwen3-4B-Instruct-2507. Response quality is assessed using the carefully designed fine-grained rubrics introduced by \citet{researchrubrics}, which measure aspects such as factual grounding and clarity, with Gemini 2.5 Pro \citep{comanici2025gemini} serving as the evaluator of model-generated outputs. Notably, ResearchRubrics covers ten diverse domains, including STEM and Historical Analysis. We report both per-domain scores and the overall average score in \Cref{fig:researchrubrics}.

As shown in \Cref{fig:researchrubrics}, CCS achieves the best overall performance on both models, outperforming not only methods that do not rely on gold supervision, but also Search-R1, which is trained with gold supervision. Interestingly, Search-R1, which is optimized to generate ground-truth answers for closed-ended questions, exhibits relatively weaker performance on this open-ended task. More specifically, on Qwen2.5-7B-Instruct, CCS achieves relative improvements of 7.92\%, 14.48\%, 17.63\%, and 9.96\% over Search-O1, Search-R1, RLIF, and CJ, respectively. On Qwen3-4B-Instruct-2507, the corresponding relative improvements are 20.08\%, 19.53\%, 26.81\%, and 7.86\%.
\section{Conclusion}
We introduced Cycle-Consistent Search (CCS), a gold-supervision-free framework for training search agents. By treating search trajectories as intermediate representations of the original question, CCS uses cycle-consistency to derive reward signals from reconstruction quality. To reduce information leakage, we incorporate information bottlenecks through final-response exclusion and named entity masking on search actions. Across multiple benchmarks, CCS achieves performance comparable to supervised methods while consistently outperforming competing methods without gold supervision. Overall, these results show that cycle-consistency provides a promising and scalable objective for training search agents in settings where gold supervision is unavailable.

\bibliographystyle{assets/plainnat}
\bibliography{paper}

@article{brown2020language,
  title={Language models are few-shot learners},
  author={Brown, Tom and Mann, Benjamin and Ryder, Nick and Subbiah, Melanie and Kaplan, Jared D and Dhariwal, Prafulla and Neelakantan, Arvind and Shyam, Pranav and Sastry, Girish and Askell, Amanda and others},
  journal={Advances in neural information processing systems},
  volume={33},
  pages={1877--1901},
  year={2020}
}

@article{chung2024scaling,
  title={Scaling instruction-finetuned language models},
  author={Chung, Hyung Won and Hou, Le and Longpre, Shayne and Zoph, Barret and Tay, Yi and Fedus, William and Li, Yunxuan and Wang, Xuezhi and Dehghani, Mostafa and Brahma, Siddhartha and others},
  journal={Journal of Machine Learning Research},
  volume={25},
  number={70},
  pages={1--53},
  year={2024}
}

@article{ppo,
  title={Proximal policy optimization algorithms},
  author={Schulman, John and Wolski, Filip and Dhariwal, Prafulla and Radford, Alec and Klimov, Oleg},
  journal={arXiv preprint arXiv:1707.06347},
  year={2017}
}

@article{grpo,
  title={Deepseekmath: Pushing the limits of mathematical reasoning in open language models},
  author={Shao, Zhihong and Wang, Peiyi and Zhu, Qihao and Xu, Runxin and Song, Junxiao and Bi, Xiao and Zhang, Haowei and Zhang, Mingchuan and Li, YK and Wu, Yang and others},
  journal={arXiv preprint arXiv:2402.03300},
  year={2024}
}

@article{search-r1,
  title={Search-r1: Training llms to reason and leverage search engines with reinforcement learning},
  author={Jin, Bowen and Zeng, Hansi and Yue, Zhenrui and Yoon, Jinsung and Arik, Sercan and Wang, Dong and Zamani, Hamed and Han, Jiawei},
  journal={arXiv preprint arXiv:2503.09516},
  year={2025}
}

@article{r1-searcher,
  title={R1-searcher: Incentivizing the search capability in llms via reinforcement learning},
  author={Song, Huatong and Jiang, Jinhao and Min, Yingqian and Chen, Jie and Chen, Zhipeng and Zhao, Wayne Xin and Fang, Lei and Wen, Ji-Rong},
  journal={arXiv preprint arXiv:2503.05592},
  year={2025}
}

@article{deepresearcher,
  title={Deepresearcher: Scaling deep research via reinforcement learning in real-world environments},
  author={Zheng, Yuxiang and Fu, Dayuan and Hu, Xiangkun and Cai, Xiaojie and Ye, Lyumanshan and Lu, Pengrui and Liu, Pengfei},
  journal={arXiv preprint arXiv:2504.03160},
  year={2025}
}

@article{ReSearch,
  title={Learning to reason with search for llms via reinforcement learning},
  author={Chen, Mingyang and Sun, Linzhuang and Li, Tianpeng and Sun, Haoze and Zhou, Yijie and Zhu, Chenzheng and Wang, Haofen and Pan, Jeff Z and Zhang, Wen and Chen, Huajun and others},
  journal={arXiv preprint arXiv:2503.19470},
  year={2025}
}

@article{search-o1,
  title={Search-o1: Agentic search-enhanced large reasoning models},
  author={Li, Xiaoxi and Dong, Guanting and Jin, Jiajie and Zhang, Yuyao and Zhou, Yujia and Zhu, Yutao and Zhang, Peitian and Dou, Zhicheng},
  journal={arXiv preprint arXiv:2501.05366},
  year={2025}
}

@article{he2016dual,
  title={Dual learning for machine translation},
  author={He, Di and Xia, Yingce and Qin, Tao and Wang, Liwei and Yu, Nenghai and Liu, Tie-Yan and Ma, Wei-Ying},
  journal={Advances in neural information processing systems},
  volume={29},
  year={2016}
}

@article{lample2017unsupervised,
  title={Unsupervised machine translation using monolingual corpora only},
  author={Lample, Guillaume and Conneau, Alexis and Denoyer, Ludovic and Ranzato, Marc'Aurelio},
  journal={arXiv preprint arXiv:1711.00043},
  year={2017}
}

@article{han2021unsupervised,
  title={Unsupervised neural machine translation with generative language models only},
  author={Han, Jesse Michael and Babuschkin, Igor and Edwards, Harrison and Neelakantan, Arvind and Xu, Tao and Polu, Stanislas and Ray, Alex and Shyam, Pranav and Ramesh, Aditya and Radford, Alec and others},
  journal={arXiv preprint arXiv:2110.05448},
  year={2021}
}

@inproceedings{cyclegan,
  title={Unpaired image-to-image translation using cycle-consistent adversarial networks},
  author={Zhu, Jun-Yan and Park, Taesung and Isola, Phillip and Efros, Alexei A},
  booktitle={Proceedings of the IEEE international conference on computer vision},
  pages={2223--2232},
  year={2017}
}

@article{liu2017unsupervised,
  title={Unsupervised image-to-image translation networks},
  author={Liu, Ming-Yu and Breuel, Thomas and Kautz, Jan},
  journal={Advances in neural information processing systems},
  volume={30},
  year={2017}
}

@inproceedings{huang2018multimodal,
  title={Multimodal unsupervised image-to-image translation},
  author={Huang, Xun and Liu, Ming-Yu and Belongie, Serge and Kautz, Jan},
  booktitle={Proceedings of the European conference on computer vision (ECCV)},
  pages={172--189},
  year={2018}
}

@article{yamada2020luke,
  title={LUKE: Deep contextualized entity representations with entity-aware self-attention},
  author={Yamada, Ikuya and Asai, Akari and Shindo, Hiroyuki and Takeda, Hideaki and Matsumoto, Yuji},
  journal={arXiv preprint arXiv:2010.01057},
  year={2020}
}

@article{sun2019ernie,
  title={Ernie: Enhanced representation through knowledge integration},
  author={Sun, Yu and Wang, Shuohuan and Li, Yukun and Feng, Shikun and Chen, Xuyi and Zhang, Han and Tian, Xin and Zhu, Danxiang and Tian, Hao and Wu, Hua},
  journal={arXiv preprint arXiv:1904.09223},
  year={2019}
}

@inproceedings{liu2020k,
  title={K-bert: Enabling language representation with knowledge graph},
  author={Liu, Weijie and Zhou, Peng and Zhao, Zhe and Wang, Zhiruo and Ju, Qi and Deng, Haotang and Wang, Ping},
  booktitle={Proceedings of the AAAI conference on artificial intelligence},
  volume={34},
  number={03},
  pages={2901--2908},
  year={2020}
}

@inproceedings{hotpotqa,
  title={HotpotQA: A dataset for diverse, explainable multi-hop question answering},
  author={Yang, Zhilin and Qi, Peng and Zhang, Saizheng and Bengio, Yoshua and Cohen, William and Salakhutdinov, Ruslan and Manning, Christopher D},
  booktitle={Proceedings of the 2018 conference on empirical methods in natural language processing},
  pages={2369--2380},
  year={2018}
}

@article{2wikimqa,
  title={Constructing a multi-hop qa dataset for comprehensive evaluation of reasoning steps},
  author={Ho, Xanh and Nguyen, Anh-Khoa Duong and Sugawara, Saku and Aizawa, Akiko},
  journal={arXiv preprint arXiv:2011.01060},
  year={2020}
}

@article{musique,
  title={MuSiQue: Multihop Questions via Single-hop Question Composition},
  author={Trivedi, Harsh and Balasubramanian, Niranjan and Khot, Tushar and Sabharwal, Ashish},
  journal={Transactions of the Association for Computational Linguistics},
  volume={10},
  pages={539--554},
  year={2022},
  publisher={MIT Press One Broadway, 12th Floor, Cambridge, Massachusetts 02142, USA~…}
}

@inproceedings{bamboogle,
  title={Measuring and narrowing the compositionality gap in language models},
  author={Press, Ofir and Zhang, Muru and Min, Sewon and Schmidt, Ludwig and Smith, Noah A and Lewis, Mike},
  booktitle={Findings of the Association for Computational Linguistics: EMNLP 2023},
  pages={5687--5711},
  year={2023}
}

@article{nq, 
title = "Natural Questions: A Benchmark for Question Answering Research", 
author = "Kwiatkowski, Tom  and Palomaki, Jennimaria  and Redfield, Olivia  and Collins, Michael  and Parikh, Ankur  and Alberti, Chris  and Epstein, Danielle  and Polosukhin, Illia  and Devlin, Jacob  and Lee, Kenton  and Toutanova, Kristina  and Jones, Llion  and Kelcey, Matthew  and Chang, Ming-Wei  and Dai, Andrew M.  and Uszkoreit, Jakob  and Le, Quoc  and Petrov, Slav", 
editor = "Lee, Lillian  and Johnson, Mark  and Roark, Brian  and Nenkova, Ani", 
journal = "Transactions of the Association for Computational Linguistics", 
volume = "7", 
year = "2019", 
address = "Cambridge, MA", 
publisher = "MIT Press", 
url = "https://aclanthology.org/Q19-1026/", 
doi = "10.1162/tacl_a_00276", 
pages = "452--466" }

@article{triviaqa,
  title={Triviaqa: A large scale distantly supervised challenge dataset for reading comprehension},
  author={Joshi, Mandar and Choi, Eunsol and Weld, Daniel S and Zettlemoyer, Luke},
  journal={arXiv preprint arXiv:1705.03551},
  year={2017}
}

@inproceedings{popqa,
  title={When not to trust language models: Investigating effectiveness of parametric and non-parametric memories},
  author={Mallen, Alex and Asai, Akari and Zhong, Victor and Das, Rajarshi and Khashabi, Daniel and Hajishirzi, Hannaneh},
  booktitle={Proceedings of the 61st Annual Meeting of the Association for Computational Linguistics (Volume 1: Long Papers)},
  pages={9802--9822},
  year={2023}
}

@article{researchrubrics,
  title={Researchrubrics: A benchmark of prompts and rubrics for evaluating deep research agents},
  author={Sharma, Manasi and Zhang, Chen Bo Calvin and Bandi, Chaithanya and Wang, Clinton and Aich, Ankit and Nghiem, Huy and Rabbani, Tahseen and Htet, Ye and Jang, Brian and Basu, Sumana and others},
  journal={arXiv preprint arXiv:2511.07685},
  year={2025}
}

@article{yang2025qwen3,
  title={Qwen3 technical report},
  author={Yang, An and Li, Anfeng and Yang, Baosong and Zhang, Beichen and Hui, Binyuan and Zheng, Bo and Yu, Bowen and Gao, Chang and Huang, Chengen and Lv, Chenxu and others},
  journal={arXiv preprint arXiv:2505.09388},
  year={2025}
}

@article{comanici2025gemini,
  title={Gemini 2.5: Pushing the frontier with advanced reasoning, multimodality, long context, and next generation agentic capabilities},
  author={Comanici, Gheorghe and Bieber, Eric and Schaekermann, Mike and Pasupat, Ice and Sachdeva, Noveen and Dhillon, Inderjit and Blistein, Marcel and Ram, Ori and Zhang, Dan and Rosen, Evan and others},
  journal={arXiv preprint arXiv:2507.06261},
  year={2025}
}

@article{qwen3embedding,
  title={Qwen3 Embedding: Advancing Text Embedding and Reranking Through Foundation Models},
  author={Zhang, Yanzhao and Li, Mingxin and Long, Dingkun and Zhang, Xin and Lin, Huan and Yang, Baosong and Xie, Pengjun and Yang, An and Liu, Dayiheng and Lin, Junyang and Huang, Fei and Zhou, Jingren},
  journal={arXiv preprint arXiv:2506.05176},
  year={2025}
}

@article{bert-base-NER,
  author    = {Jacob Devlin and
               Ming{-}Wei Chang and
               Kenton Lee and
               Kristina Toutanova},
  title     = {{BERT:} Pre-training of Deep Bidirectional Transformers for Language
               Understanding},
  journal   = {CoRR},
  volume    = {abs/1810.04805},
  year      = {2018},
  url       = {http://arxiv.org/abs/1810.04805},
  archivePrefix = {arXiv},
  eprint    = {1810.04805},
  bibsource = {dblp computer science bibliography, https://dblp.org}
}

@article{cot,
  title={Chain-of-thought prompting elicits reasoning in large language models},
  author={Wei, Jason and Wang, Xuezhi and Schuurmans, Dale and Bosma, Maarten and Xia, Fei and Chi, Ed and Le, Quoc V and Zhou, Denny and others},
  journal={Advances in neural information processing systems},
  volume={35},
  pages={24824--24837},
  year={2022}
}

@inproceedings{ircot,
  title={Interleaving retrieval with chain-of-thought reasoning for knowledge-intensive multi-step questions},
  author={Trivedi, Harsh and Balasubramanian, Niranjan and Khot, Tushar and Sabharwal, Ashish},
  booktitle={Proceedings of the 61st annual meeting of the association for computational linguistics (volume 1: long papers)},
  pages={10014--10037},
  year={2023}
}

@article{rag,
  title={Retrieval-augmented generation for knowledge-intensive nlp tasks},
  author={Lewis, Patrick and Perez, Ethan and Piktus, Aleksandra and Petroni, Fabio and Karpukhin, Vladimir and Goyal, Naman and K{\"u}ttler, Heinrich and Lewis, Mike and Yih, Wen-tau and Rockt{\"a}schel, Tim and others},
  journal={Advances in neural information processing systems},
  volume={33},
  pages={9459--9474},
  year={2020}
}

@article{rlif,
  title={Learning to reason without external rewards},
  author={Zhao, Xuandong and Kang, Zhewei and Feng, Aosong and Levine, Sergey and Song, Dawn},
  journal={arXiv preprint arXiv:2505.19590},
  year={2025}
}

@article{ttrl,
  title={Ttrl: Test-time reinforcement learning},
  author={Zuo, Yuxin and Zhang, Kaiyan and Sheng, Li and Qu, Shang and Cui, Ganqu and Zhu, Xuekai and Li, Haozhan and Zhang, Yuchen and Long, Xinwei and Hua, Ermo and others},
  journal={arXiv preprint arXiv:2504.16084},
  year={2025}
}

@inproceedings{geval,
  title={G-eval: NLG evaluation using gpt-4 with better human alignment},
  author={Liu, Yang and Iter, Dan and Xu, Yichong and Wang, Shuohang and Xu, Ruochen and Zhu, Chenguang},
  booktitle={Proceedings of the 2023 conference on empirical methods in natural language processing},
  pages={2511--2522},
  year={2023}
}

@article{borlund2003iir,
  author  = {Borlund, Pia},
  title   = {The IIR Evaluation Model: A Framework for Evaluation of Interactive Information Retrieval Systems},
  journal = {Information Research},
  volume  = {8},
  number  = {3},
  year    = {2003},
  note    = {Paper No. 152},
  url     = {http://informationr.net/ir/8-3/paper152.html}
}

@article{kelly2009methods,
  title={Methods for evaluating interactive information retrieval systems with users},
  author={Kelly, Diane},
  journal={Foundations and Trends{\textregistered} in Information Retrieval},
  volume={3},
  number={1--2},
  pages={1--224},
  year={2009},
  publisher={Emerald Publishing Limited}
}

@inproceedings{carterette2011system,
  title={System effectiveness, user models, and user utility: a conceptual framework for investigation},
  author={Carterette, Ben},
  booktitle={Proceedings of the 34th international ACM SIGIR conference on Research and development in information retrieval},
  pages={903--912},
  year={2011}
}

@inproceedings{react,
  title={React: Synergizing reasoning and acting in language models},
  author={Yao, Shunyu and Zhao, Jeffrey and Yu, Dian and Du, Nan and Shafran, Izhak and Narasimhan, Karthik R and Cao, Yuan},
  booktitle={The eleventh international conference on learning representations},
  year={2022}
}

@article{e5,
  title={Text embeddings by weakly-supervised contrastive pre-training},
  author={Wang, Liang and Yang, Nan and Huang, Xiaolong and Jiao, Binxing and Yang, Linjun and Jiang, Daxin and Majumder, Rangan and Wei, Furu},
  journal={arXiv preprint arXiv:2212.03533},
  year={2022}
}

@inproceedings{wikidump,
  title={Dense Passage Retrieval for Open-Domain Question Answering.},
  author={Karpukhin, Vladimir and Oguz, Barlas and Min, Sewon and Lewis, Patrick SH and Wu, Ledell and Edunov, Sergey and Chen, Danqi and Yih, Wen-tau},
  booktitle={EMNLP (1)},
  pages={6769--6781},
  year={2020}
}

@inproceedings{guu2020retrieval,
  title={Retrieval augmented language model pre-training},
  author={Guu, Kelvin and Lee, Kenton and Tung, Zora and Pasupat, Panupong and Chang, Mingwei},
  booktitle={International conference on machine learning},
  pages={3929--3938},
  year={2020},
  organization={PMLR}
}

@inproceedings{vllm,
  title={Efficient Memory Management for Large Language Model Serving with PagedAttention},
  author={Woosuk Kwon and Zhuohan Li and Siyuan Zhuang and Ying Sheng and Lianmin Zheng and Cody Hao Yu and Joseph E. Gonzalez and Hao Zhang and Ion Stoica},
  booktitle={Proceedings of the ACM SIGOPS 29th Symposium on Operating Systems Principles},
  year={2023}
}

@inproceedings{FlashRAG,
  author       = {Jiajie Jin and
                  Yutao Zhu and
                  Zhicheng Dou and
                  Guanting Dong and
                  Xinyu Yang and
                  Chenghao Zhang and
                  Tong Zhao and
                  Zhao Yang and
                  Ji{-}Rong Wen},
  editor       = {Guodong Long and
                  Michale Blumestein and
                  Yi Chang and
                  Liane Lewin{-}Eytan and
                  Zi Helen Huang and
                  Elad Yom{-}Tov},
  title        = {FlashRAG: {A} Modular Toolkit for Efficient Retrieval-Augmented Generation
                  Research},
  booktitle    = {Companion Proceedings of the {ACM} on Web Conference 2025, {WWW} 2025,
                  Sydney, NSW, Australia, 28 April 2025 - 2 May 2025},
  pages        = {737--740},
  publisher    = {{ACM}},
  year         = {2025},
  url          = {https://doi.org/10.1145/3701716.3715313},
  doi          = {10.1145/3701716.3715313}
}

@software{axolotl,
  title = {Axolotl: Open Source LLM Post-Training},
  author = {{Axolotl maintainers and contributors}},
  url = {https://github.com/axolotl-ai-cloud/axolotl},
  license = {Apache-2.0},
  year = {2023}
}

@inproceedings{lee-etal-2019-latent,
    title = "Latent Retrieval for Weakly Supervised Open Domain Question Answering",
    author = "Lee, Kenton  and
      Chang, Ming-Wei  and
      Toutanova, Kristina",
    editor = "Korhonen, Anna  and
      Traum, David  and
      M{\`a}rquez, Llu{\'i}s",
    booktitle = "Proceedings of the 57th Annual Meeting of the Association for Computational Linguistics",
    month = jul,
    year = "2019",
    address = "Florence, Italy",
    publisher = "Association for Computational Linguistics",
    url = "https://aclanthology.org/P19-1612/",
    doi = "10.18653/v1/P19-1612",
    pages = "6086--6096"
}

@inproceedings{karpukhin-etal-2020-dense,
    title = "Dense Passage Retrieval for Open-Domain Question Answering",
    author = "Karpukhin, Vladimir  and
      Oguz, Barlas  and
      Min, Sewon  and
      Lewis, Patrick  and
      Wu, Ledell  and
      Edunov, Sergey  and
      Chen, Danqi  and
      Yih, Wen-tau",
    editor = "Webber, Bonnie  and
      Cohn, Trevor  and
      He, Yulan  and
      Liu, Yang",
    booktitle = "Proceedings of the 2020 Conference on Empirical Methods in Natural Language Processing (EMNLP)",
    month = nov,
    year = "2020",
    address = "Online",
    publisher = "Association for Computational Linguistics",
    url = "https://aclanthology.org/2020.emnlp-main.550/",
    doi = "10.18653/v1/2020.emnlp-main.550",
    pages = "6769--6781"
}

\clearpage
\newpage
\beginappendix

\appendix
\section{Additional Results}

\subsection{Training Dynamics}
\begin{figure}[htbp]
\centering

\begin{subfigure}{\textwidth}
    \centering
    \includegraphics[width=0.3\textwidth]{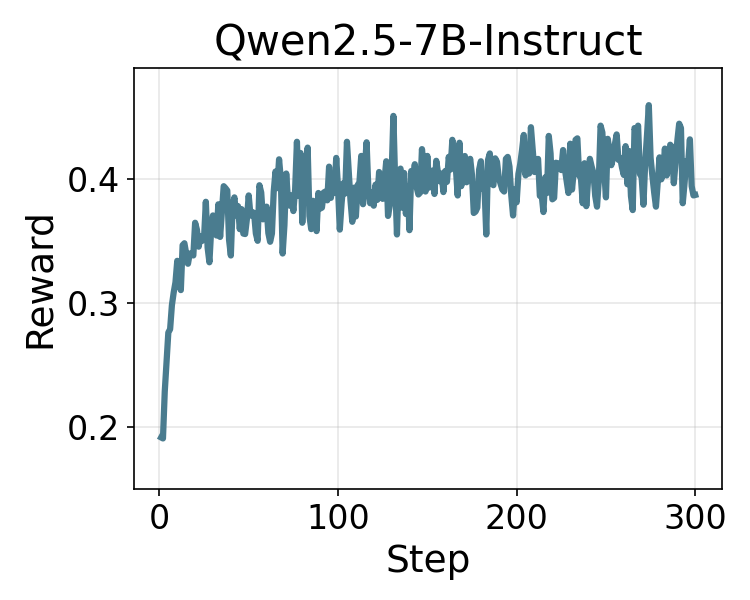}
    \hfill
    \includegraphics[width=0.3\textwidth]{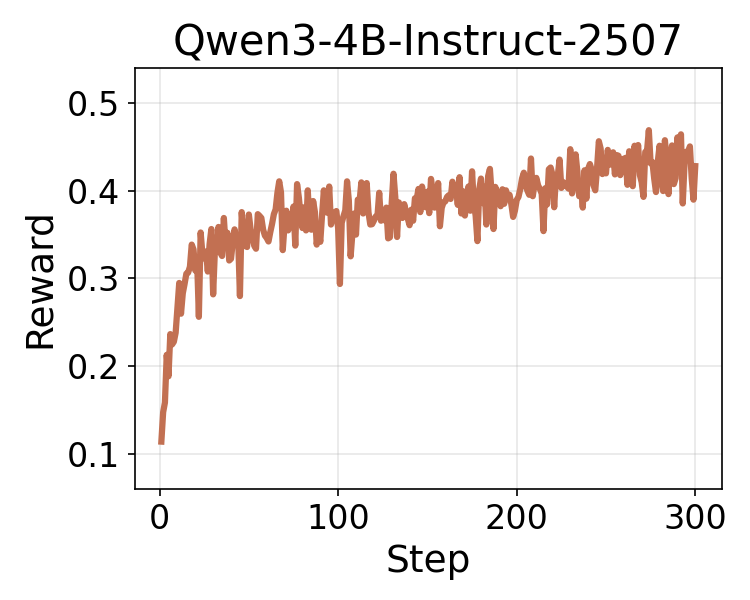}
    \hfill
    \includegraphics[width=0.3\textwidth]{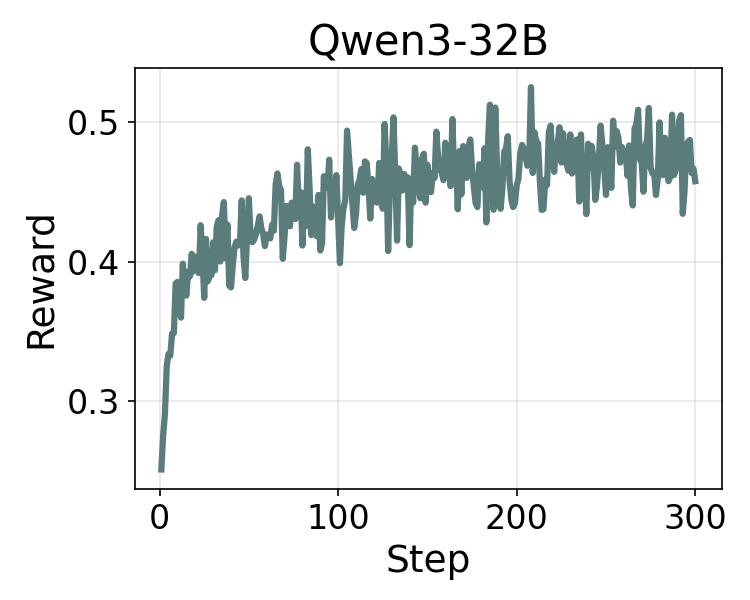}
    \caption{Search-R1}
\end{subfigure}

\vspace{0.5em}

\begin{subfigure}{\textwidth}
    \centering
    \includegraphics[width=0.3\textwidth]{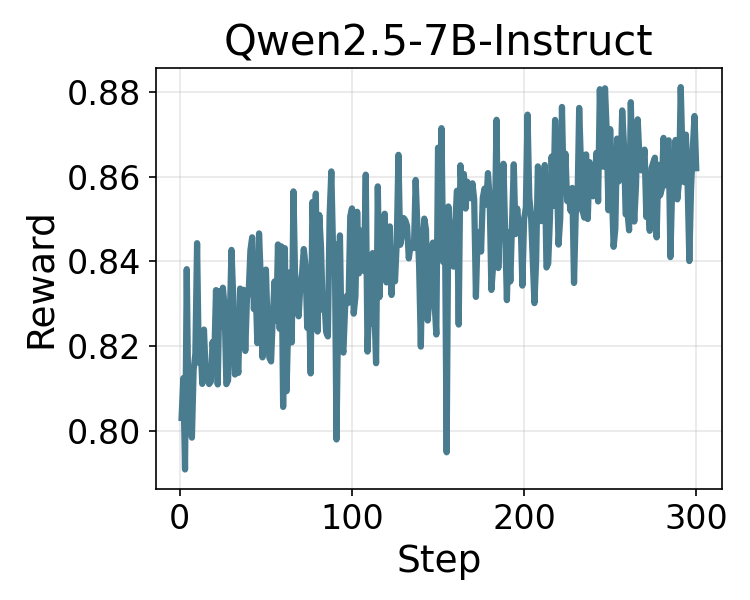}
    \hfill
    \includegraphics[width=0.3\textwidth]{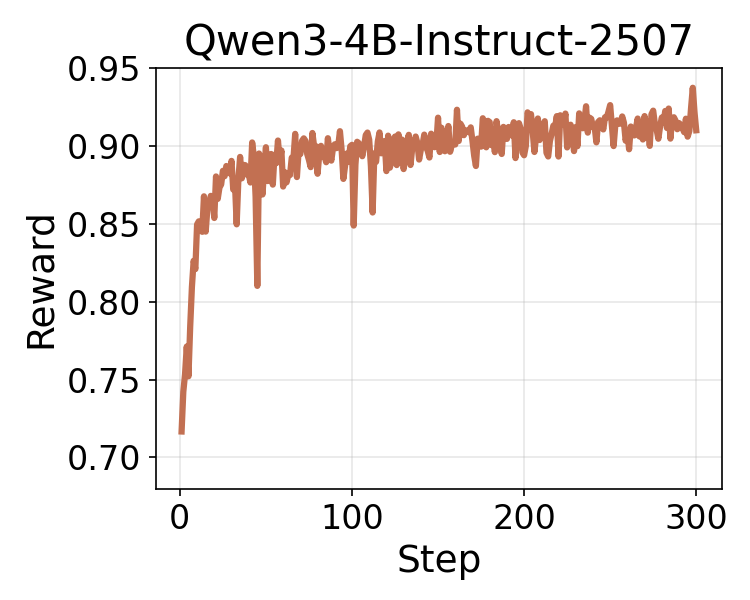}
    \hfill
    \includegraphics[width=0.3\textwidth]{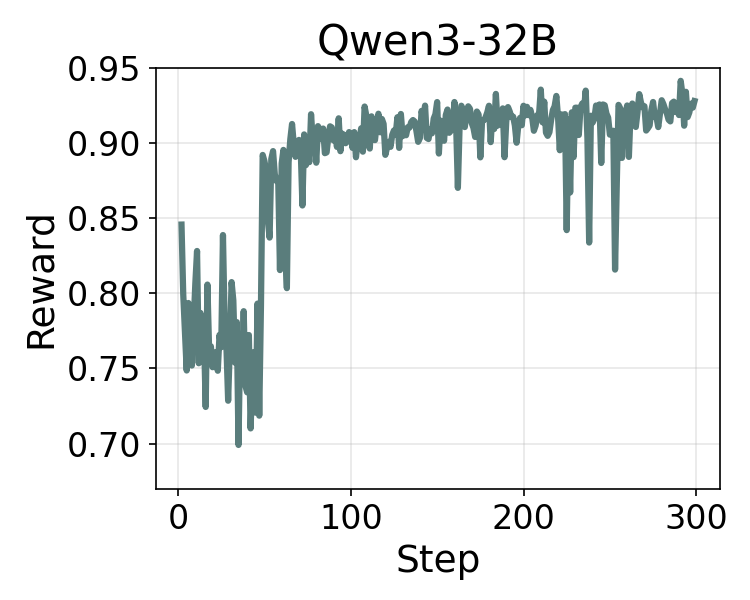}
    \caption{Constitutional Judge (CJ)}
\end{subfigure}

\vspace{0.5em}

\begin{subfigure}{\textwidth}
    \centering
    \includegraphics[width=0.3\textwidth]{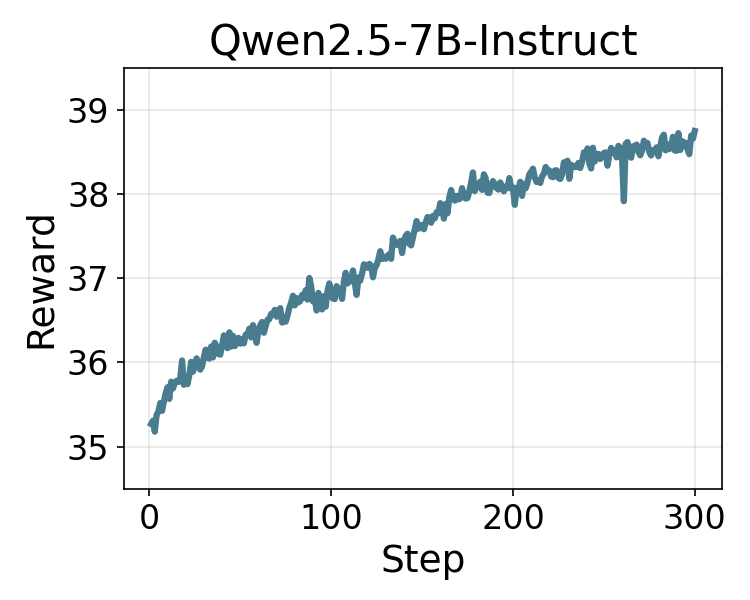}
    \hfill
    \includegraphics[width=0.3\textwidth]{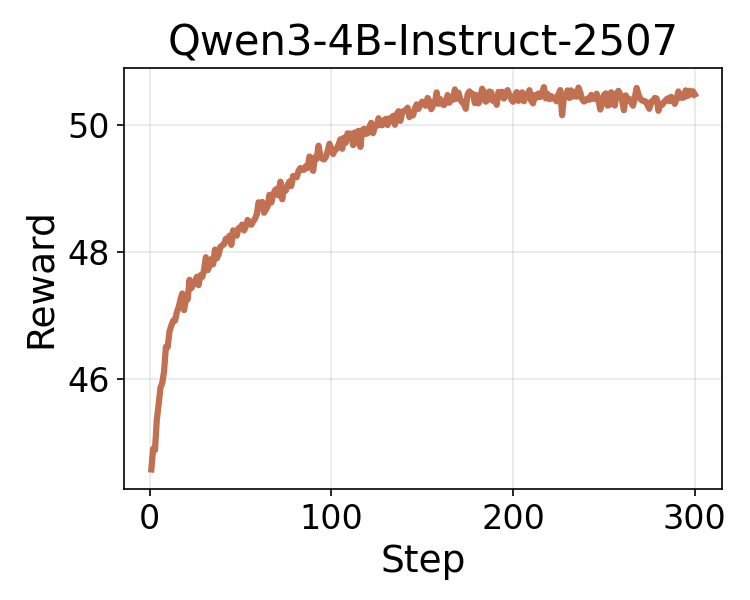}
    \hfill
    \includegraphics[width=0.3\textwidth]{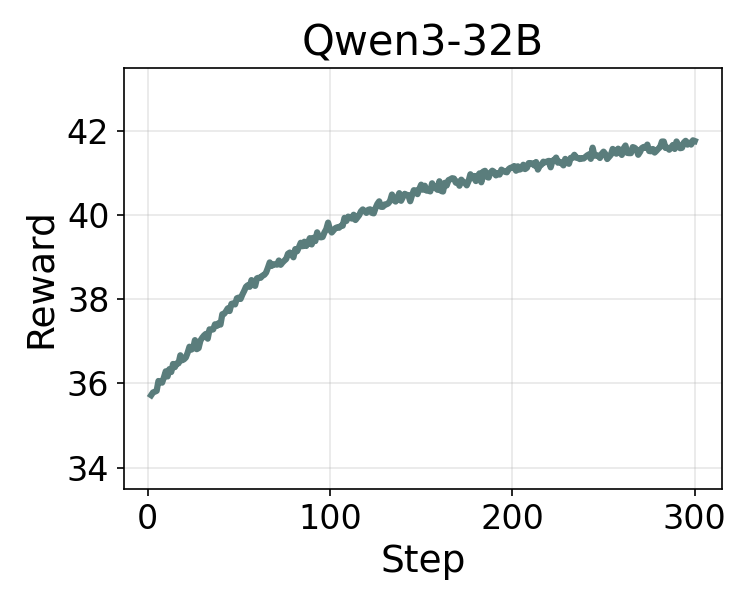}
    \caption{RLIF}
\end{subfigure}

\vspace{0.5em}

\begin{subfigure}{\textwidth}
    \centering
    \includegraphics[width=0.3\textwidth]{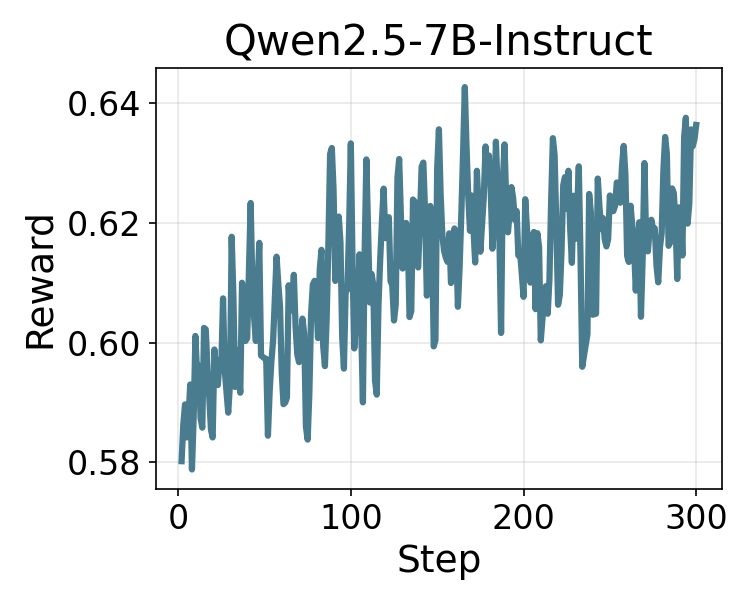}
    \hfill
    \includegraphics[width=0.3\textwidth]{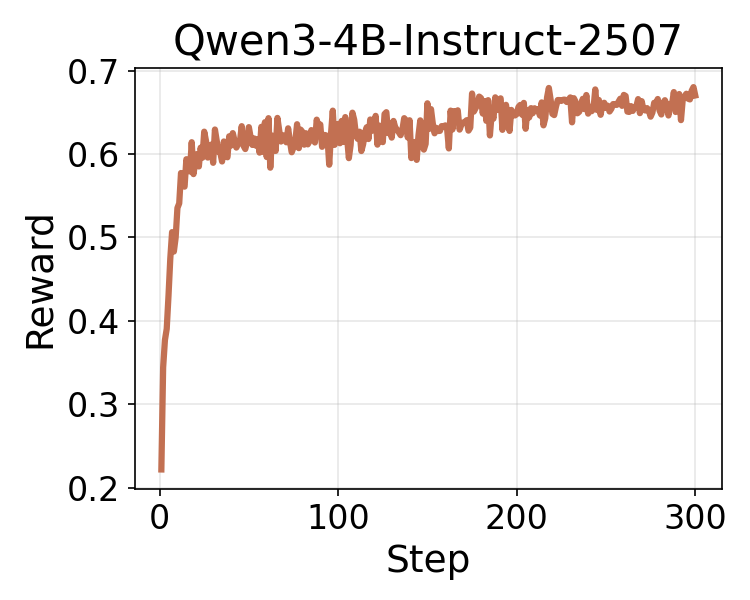}
    \hfill
    \includegraphics[width=0.3\textwidth]{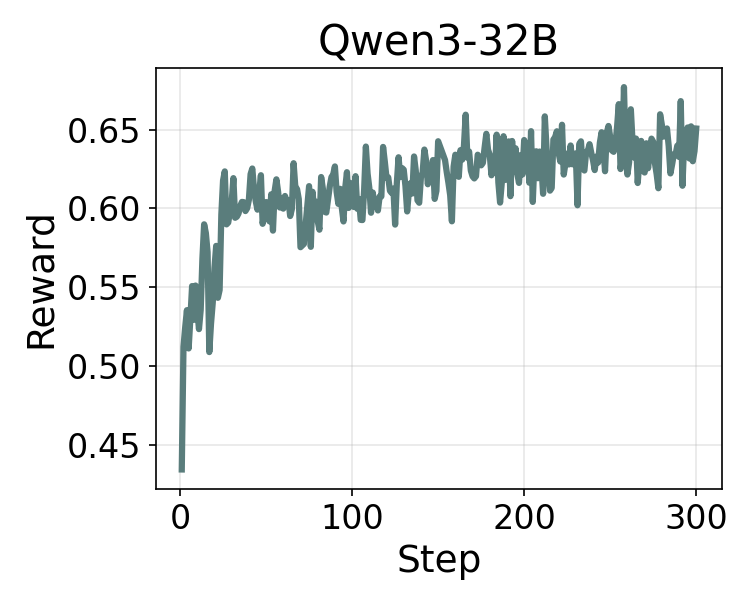}
    \caption{CCS}
\end{subfigure}

\caption{\textbf{Training Reward during RL.}}
\label{fig:rl_train_reward}

\end{figure}

\begin{figure}[htbp]
\centering

\begin{subfigure}{0.32\textwidth}
\centering
\includegraphics[width=\linewidth]{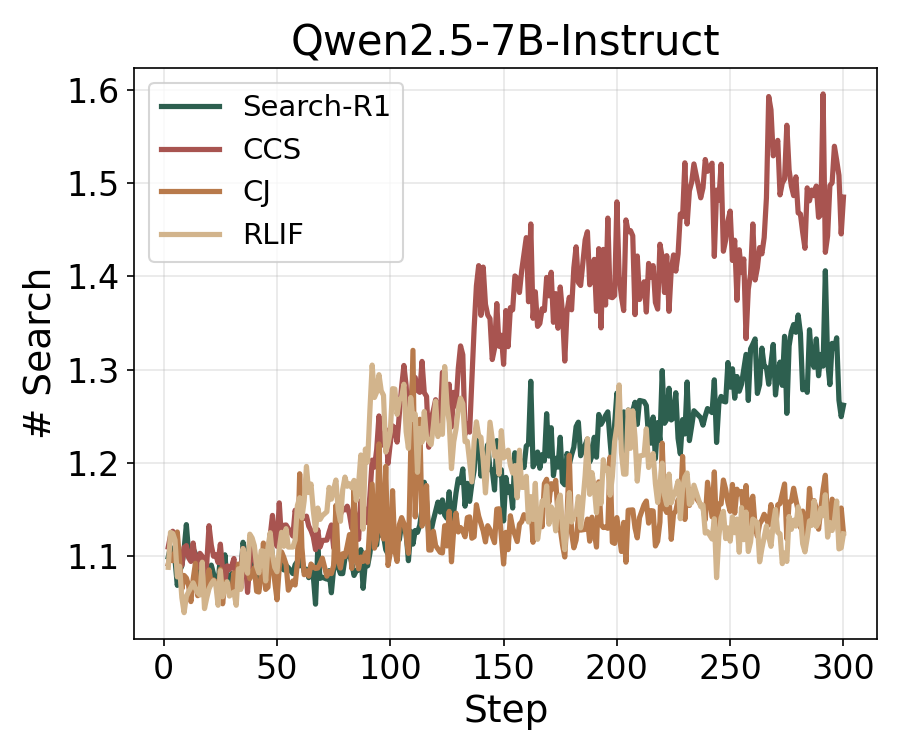}
\end{subfigure}
\hfill
\begin{subfigure}{0.32\textwidth}
\centering
\includegraphics[width=\linewidth]{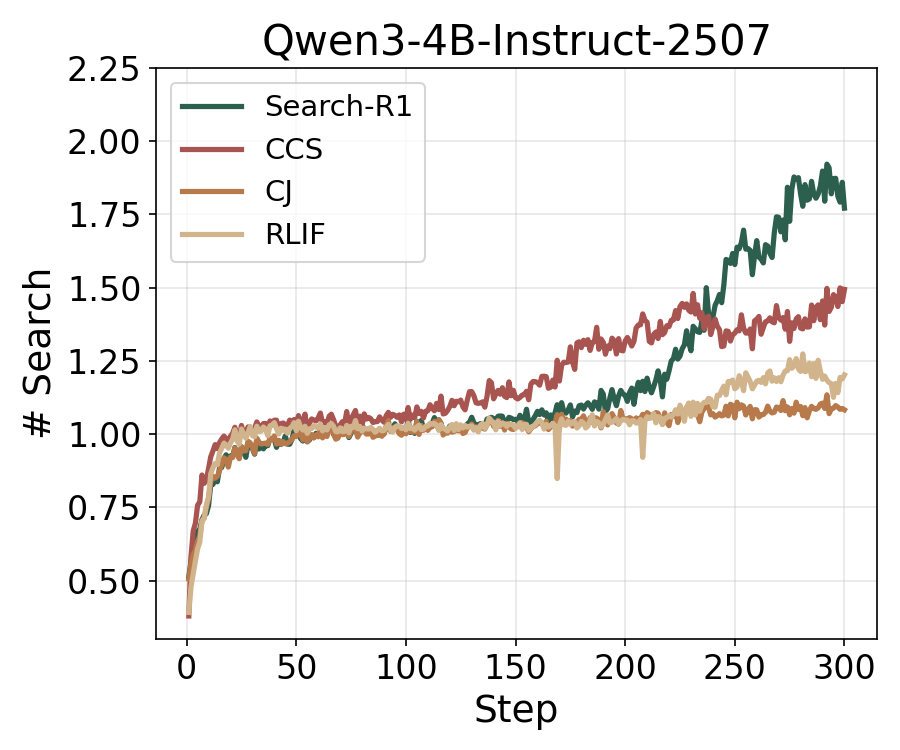}
\end{subfigure}
\hfill
\begin{subfigure}{0.32\textwidth}
\centering
\includegraphics[width=\linewidth]{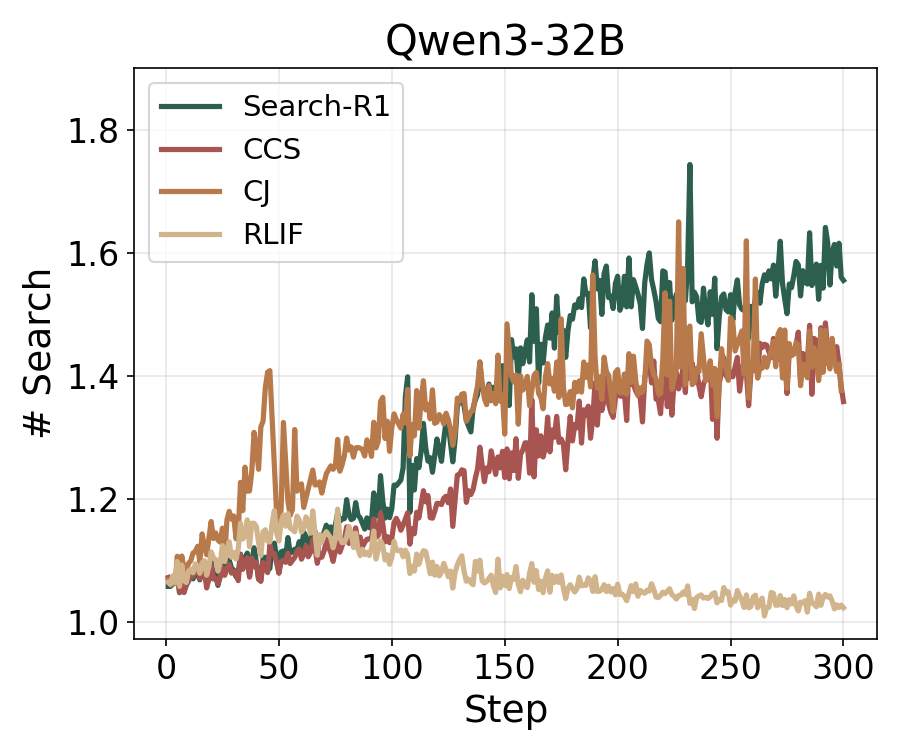}
\end{subfigure}

\caption{\textbf{Average number of search during RL.}}
\label{fig:rl_avg_num_search}

\end{figure}

We plot the training rewards of Search-R1, Constitutional Judge (CJ), RLIF, and CCS in \Cref{fig:rl_train_reward}. Since each method uses a different reward function and therefore operates on a different scale, we present them in separate plots. Additionally, the average number of searches is plotted in \Cref{fig:rl_avg_num_search}.

\section{Implementation Details}
\label{appendix:implementaion_details}

\subsection{Hyperparameters}
\label{appendix:hyperparams}
We conduct experiments using three policy models: Qwen2.5-7B-Instruct, Qwen3-4B-Instruct-2507, and Qwen3-32B \citep{yang2025qwen3}. We use Gemini 2.5 Flash \citep{comanici2025gemini} as both the evaluator and reconstructor. We use Qwen3-Embedding-4B \citep{qwen3embedding} to compute the semantic similarity between $q$ and $\hat{q}$ in \Cref{eq:embedding_sim}, and bert-base-NER \citep{bert-base-NER} to mask search actions.

Following \citet{search-r1}, we train all fine-tuning-based baselines, including SFT and RL methods, on the merged training sets of NQ and HotpotQA. We evaluate the models on the test or validation splits of seven datasets to assess both in-domain and out-of-domain generalization. For evaluation, we use Gemini 2.5 Flash as the evaluator with the prompt described in \Cref{appendix:evaluator_prompt}. The evaluation temperature is set to 0.1.

For RL-based methods, we set the policy learning rate to $1\times10^{-6}$ and sample five responses per prompt. Training is performed for 300 steps using the FSDP2 strategy, with gradient checkpointing enabled to improve GPU memory efficiency.

All training runs are conducted on four nodes, each equipped with eight H100 GPUs. We use a total batch size of 512, a mini-batch size of 256, and a micro-batch size of 64 for Qwen2.5-7B-Instruct and Qwen3-4B-Instruct-2507. For Qwen3-32B, we use a micro-batch size of 32 due to memory constraints. The maximum context length is set to 30,000 tokens. The maximum response length for each step is 512 tokens, and the maximum search response length for each step is 8,192 tokens.

For efficient rollout generation, we use vLLM \citep{vllm} with tensor parallelism of size 1 for Qwen2.5-7B-Instruct and Qwen3-4B-Instruct-2507, and size 2 for Qwen3-32B. During rollout sampling, the temperature is set to 1.0 and top-$p$ is set to 1.0. The KL divergence regularization coefficient $\beta$ and the clip ratio $\epsilon$ are set to 0.01 and 0.2, respectively. The maximum action budget is set to 4, and up to 10 snippets are retrieved from the search engine.

\newpage
\subsection{Evaluation}
\label{appendix:evaluator_prompt}

\begin{promptbox}[]
Your job is to look at a question, a gold target, and a predicted answer, and then assign a grade of either ["CORRECT", "INCORRECT", "NOT_ATTEMPTED"].
First, I will give examples of each grade, and then you will grade a new example.


The following are examples of CORRECT predicted answers.
```
Question: What are the names of Barack Obama's children?
Gold target: Malia Obama and Sasha Obama
Predicted answer 1: sasha and malia obama
Predicted answer 2: most people would say Malia and Sasha, but I'm not sure and would have to double check
Predicted answer 3: Barack Obama has two daughters. Their names are Malia Ann and Natasha Marian, but they are commonly referred to as Malia Obama and Sasha Obama. Malia was born on July 4, 1998, and Sasha was born on June 10, 2001.
```
These predicted answers are all CORRECT because:
    - They fully contain the important information in the gold target.
    - They do not contain any information that contradicts the gold target.
    - Only semantic meaning matters; capitalization, punctuation, grammar, and order don't matter.
    - Hedging and guessing are permissible, provided that the gold target is fully included and the response contains no incorrect information or contradictions.


The following are examples of INCORRECT predicted answers.
```
Question: What are the names of Barack Obama's children?
Gold target: Malia and Sasha
Predicted answer 1: Malia.
Predicted answer 2: Malia, Sasha, and Susan.
Predicted answer 3: Barack Obama does not have any children.
Predicted answer 4: I think it's either Malia and Sasha. Or it could be Malia and Jackie. Or it could be Joey and Malia.
Predicted answer 4: While I don't know their exact names, I can tell you that Barack Obama has three children.
Predicted answer 5: It's possible you may mean Betsy and Olivia. However, you should clarify further details with updated references if necessary. Is that the correct answer?
Predicted answer 6: It may be the case that Obama's child is named James. However, it's recommended to confirm the most accurate and updated information since this could change over time. This model may not always reflect the most current information.
```
These predicted answers are all INCORRECT because:
    - A factual statement in the answer contradicts the gold target. Incorrect statements that have some hedging (e.g., "it is possible that", "although i'm not sure, i think") are also considered incorrect.


The following are examples of NOT_ATTEMPTED predicted answers.
```
Question: What are the names of Barack Obama's children?
Gold target: Malia and Sasha
Predicted answer 1: I don't know.
Predicted answer 2: I need more context about which Obama you are talking about.
Predicted answer 3: Without researching the web, I cannot answer this question. However, I can tell you that Barack Obama has two children.
Predicted answer 4: Barack Obama has two children. I know that one of them is Malia, but I'm not sure about the other one.
```
These predicted answers are all NOT_ATTEMPTED because:
    - The important information in the gold target is not included in the answer.
    - No statements in the answer contradict the gold target.


Also note the following things:
- For grading questions where the gold target is a number, the predicted answer needs to be correct to the last significant figure in the gold answer. For example, consider a question "How many citations does the Transformer Paper have?" with gold target "120k".
    - Predicted answers "120k", "124k", and 115k" are all CORRECT.
    - Predicted answers "100k" and "113k" are INCORRECT.
    - Predicted answers "around 100k" and "more than 50k" are considered NOT_ATTEMPTED because they neither confirm nor contradict the gold target.
- The gold target may contain more information than the question. In such cases, the predicted answer only needs to contain the information that is in the question.
    - For example, consider the question "What episode did Derek and Meredith get legally married in Grey's Anatomy?" with gold target "Season 7, Episode 20: White Wedding". Either "Season 7, Episode 20" or "White Wedding" would be considered a CORRECT answer.
- Do not punish predicted answers if they omit information that would be clearly inferred from the question.
    - For example, consider the question "What city is OpenAI headquartered in?" and the gold target "San Francisco, California". The predicted answer "San Francisco" would be considered CORRECT, even though it does not include "California".
    - Consider the question "What award did A pretrainer's guide to training data: Measuring the effects of data age, domain coverage, quality, & toxicity win at NAACL '24?", the gold target is "Outstanding Paper Award". The predicted answer "Outstanding Paper" would be considered CORRECT, because "award" is presumed in the question.
    - For the question "What is the height of Jason Wei in meters?", the gold target is "1.73 m". The predicted answer "1.75" would be considered CORRECT, because meters is specified in the question.
    - For the question "What is the name of Barack Obama's wife?", the gold target is "Michelle Obama". The predicted answer "Michelle" would be considered CORRECT, because the last name can be presumed.
- Do not punish for typos in people's name if it's clearly the same name.
    - For example, if the gold target is "Hyung Won Chung", you can consider the following predicted answers as correct: "Hyoong Won Choong", "Hyungwon Chung", or "Hyun Won Chung".


Here is a new example. Simply reply with either CORRECT, INCORRECT, NOT ATTEMPTED. Don't apologize or correct yourself if there was a mistake; we are just trying to grade the answer.
```
Question: {question}
Gold target: {answer}
Predicted answer: {predicted_answer}
```

Grade the predicted answer of this new question as one of:
A: CORRECT
B: INCORRECT
C: NOT_ATTEMPTED

Just return the letters "A", "B", or "C", with no text around it.
\end{promptbox}

\subsection{Reconstruction}
\begin{promptbox}
You are an expert in information recovery and intent inference. Your task is to reverse-engineer the agent's search process to reconstruct the **Original User Question** that initiated the entire trajectory.

### 1. Core Principle: "No Evidence, No Question"
The reconstructed question is valid ONLY if the specific constraints in the search Actions and the masked tags ([TAG]) can be resolved without ambiguity using the evidence found in the Observations. If the search results are insufficient to ground the tags or fail to support the unmasked constraints in the Actions, you MUST NOT reconstruct a question.

### 2. Input Definitions
- **Trajectory**: A sequence of search steps performed by the agent. Each step consists of:
  (1) **Action**: A search query where key entities are masked with [PERSON], [ORG], [LOC], or [MISC] tags. (Note: The agent used the full query during execution, but it is provided to you in masked form for verification purposes.)
  (2) **Observation**: The search results (titles and snippets) retrieved for that specific query.

### 3. Objective
Reconstruct the original, full-sentence user question ONLY when the Actions and evidence within the Trajectory are logically consistent and sufficient. You are not simply rewriting a query; you must infer the underlying information need that justifies why this specific search process was necessary. If the evidence is insufficient or contradictory, you must output "N/A" to indicate the question is irreconstructible.

### 4. Instructions
1. **Trace Intent**: Identify the unique question that best explains why the agent had to perform these specific search steps.
2. **Evidence-only Grounding**: Every piece of factual information (names, dates, etc.) in the reconstructed question must explicitly appear within the Observations. Do not use your internal pre-trained knowledge to fill in gaps.
3. **Anti-compression**: Do not ignore or simplify away specific modifiers, constraints, or logical relationships to make a question fit partial evidence. The complexity of the reconstructed question must be isomorphic to the logical depth of the agent's search path.
4. **Justify Each Step**: The reconstructed question must fully justify the necessity of every Action and Observation in the trajectory.

### 5. N/A Conditions (Output ONLY "N/A" if any apply)
1. **Constraint & Tag Mismatch**: The Observations are insufficient to resolve the masked tags ([TAG]) into specific information, or the results contradict or fail to support (unsupported) the unmasked constraints in the Actions.
2. **Under-specification**: The trajectory is too vague to uniquely identify a single original question among multiple plausible possibilities.
3. **Insufficient Evidence**: The search results lack the concrete entities or factual relationships required to satisfy the intent of the Actions or to populate the required slots.

### 6. Output Format
- Output ONLY the reconstructed question string or "N/A".
- No preface, no explanation, and no concluding remarks.
\end{promptbox}

\subsection{Baselines}


\paragraph{RAG}
We use a standard retrieve-then-read RAG framework implemented with FlashRAG's SequentialPipeline \citep{FlashRAG}. Relevant documents are retrieved from the Wikipedia DPR corpus \citep{wikidump} using E5 \citep{e5} embeddings, after which the model generates an answer conditioned on the retrieved documents. The retrieval top-$k$ is set to 10, consistent with the other methods.

\paragraph{IRCoT}
We use an IRCoT \citep{ircot} baseline implemented with FlashRAG's IRCOTPipeline \citep{FlashRAG}. Unlike standard RAG, which performs a single retrieval step before generating an answer, IRCoT follows a multi-hop reasoning process that iteratively generates a chain-of-thought step, retrieves additional documents based on the generated thought, and then produces the next reasoning step. All other hyperparameters are set identically to those of the RAG baseline.


\paragraph{SFT}
We fine-tune the model with supervised fine-tuning in the QLoRA setting using the axolotl \citep{axolotl} framework. The model is loaded in 4-bit quantized form to substantially reduce GPU memory usage, and adapters (\texttt{lora\_r}=32, \texttt{lora\_alpha}=64, \texttt{lora\_dropout}=0.05) are applied to all linear layers. We reserve 10\% of the data as a validation set and limit the sequence length to 2048 tokens. Sample packing is enabled to pack multiple samples into a single sequence, thereby improving GPU utilization efficiency. We employ FSDP, and further reduce GPU memory overhead through CPU parameter offloading. Training is conducted for one epoch with a micro-batch size of 1 and gradient accumulation over 4 steps, resulting in an effective batch size of 4. We use \texttt{adamw\_torch\_fused} as the optimizer and a cosine learning rate scheduler, with an initial learning rate of ($2\times10^{-4}$) and a warmup ratio of 10\%. To maximize memory efficiency, we enable bf16 mixed-precision training and gradient checkpointing.

\paragraph{TTRL}
Following \citet{ttrl}, we train the model on each evaluation dataset using reinforcement learning with a majority-voting reward function. Training is conducted for 300 steps. All other hyperparameters are identical to those described in \Cref{appendix:hyperparams}. \Cref{fig:ttrl_train_reward} shows the training reward for TTRL.
\begin{figure}[htbp]
\centering
\includegraphics[width=1\linewidth]{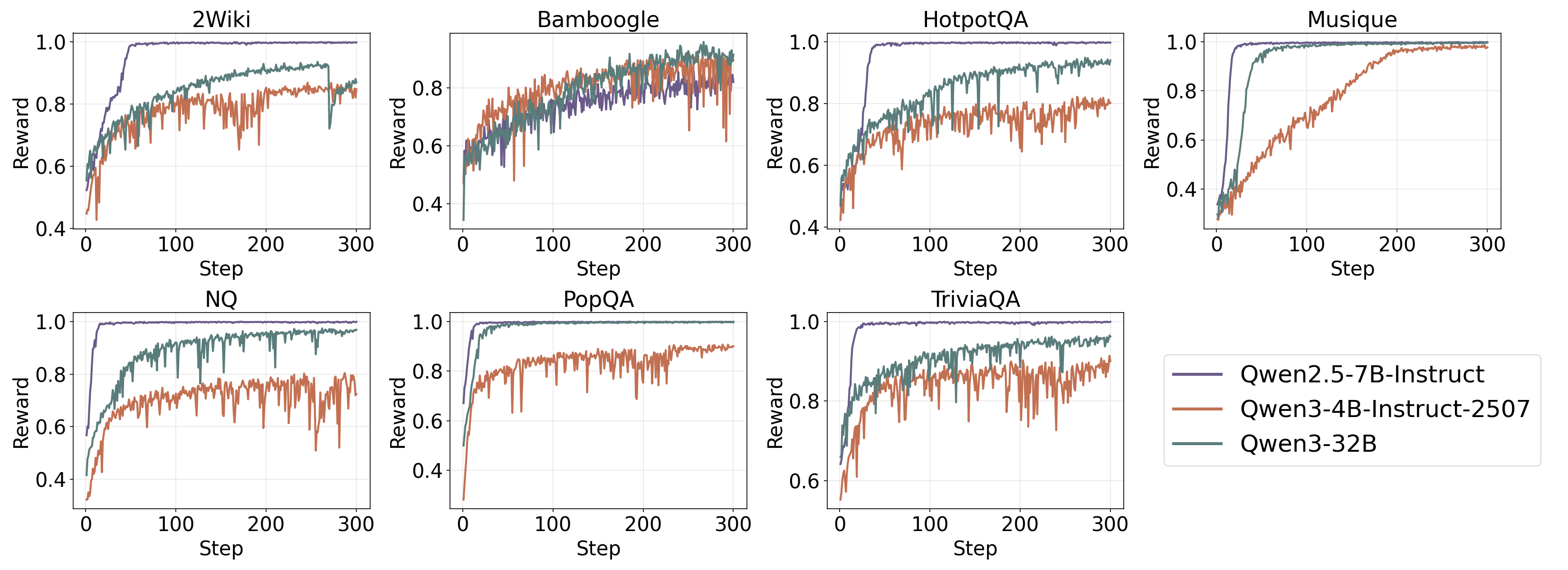}
\caption{\textbf{Training reward for TTRL.}}
\label{fig:ttrl_train_reward}
\end{figure}



\paragraph{Constitutional Judge (CJ)}
Drawing on prior work in the interactive information retrieval literature \citep{borlund2003iir, kelly2009methods, carterette2011system}, we identify three core dimensions for evaluating multi-hop interactive information retrieval systems: \textit{Response Quality}, which assesses whether the final response addresses the user's underlying information need, is factually accurate, and is appropriately scoped; \textit{Search Progression}, which evaluates whether the agent's queries develop coherently and appropriately across successive steps; and \textit{Evidence Synthesis}, which measures whether information gathered throughout the search process is accurately integrated and fully reflected in the final response. Based on these dimensions, we design the prompt for the judge model. All other hyperparameters are identical to those described in \Cref{appendix:hyperparams}.

\paragraph{RLIF}
We compute the self-certainty metric for the final response in the entire search trajectory, as defined in Equation (2) of \citet{rlif}, and use it as the reward. All other hyperparameters are identical to those described in \Cref{appendix:hyperparams}.

\section{Limitations and Future Work}
CCS demonstrates that cycle-consistency can serve as an effective training signal for search agents without gold supervision, while also opening several avenues for future work. A practical limitation of the current framework is the need for an additional reconstruction model to compute rewards. Although this overhead is small relative to the overall cost of RL training, improving the efficiency of this component or integrating it more tightly into policy optimization would further strengthen the scalability of the approach. Moreover, our formulation is developed in the context of knowledge-intensive question answering, where search trajectories can often be viewed as informative representations of user intent. Extending this idea to broader agentic settings remains an important direction for future work.


\end{document}